\documentclass{article}

\usepackage{microtype}
\usepackage{graphicx}
\usepackage{subcaption}
\usepackage{booktabs} 

\usepackage{hyperref}

\usepackage[preprint]{icml2026}

\usepackage{amsmath}
\usepackage{amssymb}
\usepackage{mathtools}
\usepackage{amsthm}
\usepackage{mathtools}
\usepackage{tcolorbox}
\usepackage{booktabs}
\usepackage{enumitem}
\usepackage{multirow}
\usepackage{multicol}
\usepackage{soul}
\usepackage{xcolor,colortbl}
\usepackage{changepage,threeparttable}
\usepackage{tabularx}
\usepackage{pifont}        
\usepackage{wrapfig}

\newcommand{\xmark}{\ding{55}}
\newcommand{\cmark}{\ding{51}}

\definecolor{myblue}{HTML}{4A90E2}
\definecolor{mygreen}{HTML}{50C878}
\definecolor{myred}{HTML}{D9534F}
\definecolor{mygray}{rgb}{0.5,0.5,0.5}
\definecolor{myorange}{HTML}{FF7A33}

\usepackage[capitalize,noabbrev]{cleveref}

\theoremstyle{plain}

\theoremstyle{definition}

\theoremstyle{remark}

\usepackage[textsize=tiny]{todonotes}

\icmltitlerunning{The Single-Multi Evolution Loop for Self-Improving  Model Collaboration Systems}

\begin{document}

\twocolumn[

  \icmltitle{The Single-Multi Evolution Loop for Self-Improving \\ Model Collaboration Systems}




\begin{icmlauthorlist}
    \icmlauthor{Shangbin Feng}{uw}
    \icmlauthor{Kishan Panaganti}{t}
    \icmlauthor{Yulia Tsvetkov}{uw}
    \icmlauthor{Wenhao Yu}{t}
  \end{icmlauthorlist}

  \icmlaffiliation{uw}{University of Washington}
  \icmlaffiliation{t}{Tencent AI Seattle Lab}

  \icmlcorrespondingauthor{Shangbin Feng}{shangbin@cs.washington.edu}


  \vskip 0.3in
]



\printAffiliationsAndNotice{}  

\begin{abstract}
  Model collaboration --- systems where multiple language models (LMs) collaborate --- combines the strengths of diverse models with cost in loading multiple LMs. We improve efficiency while preserving the strengths of collaboration by \emph{distilling collaborative patterns into a single model}, where the model is trained on the outputs of the model collaboration system. At inference time, only the distilled model is employed: it imitates the collaboration while only incurring the cost of a single model. Furthermore, we propose the \emph{single-multi evolution loop}: multiple LMs collaborate, each distills from the collaborative outputs, and these post-distillation improved LMs collaborate again, forming a collective evolution ecosystem where models evolve and self-improve by interacting with an environment of other models. Extensive experiments with 7 collaboration strategies and 15 tasks (QA, reasoning, factuality, etc.) demonstrate that: 1) individual models improve by 8.0\% on average, absorbing the strengths of collaboration while reducing the cost to a single model; 2) the collaboration also benefits from the stronger and more synergistic LMs after distillation, improving over initial systems without evolution by 14.9\% on average. Analysis reveals that the single-multi evolution loop outperforms various existing evolutionary AI methods, is compatible with diverse model/collaboration/distillation settings, and helps solve problems where the initial model/system struggles to.\footnotemark[2]\footnotetext[2]{Code and data are available at \href{https://github.com/BunsenFeng/moco_distill}{https://github.com/BunsenFeng/moco\_distill}}
\end{abstract}

\section{Introduction}
Advancing beyond a single monolithic language model (LM), an increasing line of research focuses on \emph{model collaboration} \citep{feng2025one}, where multiple LMs collaborate and complement each other: multiple LMs ``talk'' and debate with each other to divide and conquer complex problems \citep{du2023improving, jiang2025sparta}, multiple LMs are aggregated through routing systems so user queries are routed to the most fitting model \citep{ongroutellm, fenggraphrouter}, multiple LMs exchange information in the logits \citep{liutuning, shen2024learning} or model parameter \citep{huanglorahub, feng2025model} space for collective generation. These methods and algorithms combine and leverage the strengths of multiple LMs, with a caveat of increased cost: the total parameter size is (often) larger, as multiple models need to be loaded at the same time; the active parameter size is also (often) larger, especially for inference-time approaches where multiple models are all used for generation. How to reduce the cost of these modular, compositional methods and systems remains an open research question.

We propose to improve the efficiency of multi-LM systems while preserving the benefits of collaboration through \emph{distillation}: multiple LMs first collaborate to generate outputs, and we then train a single model on the output texts/logits of the collaborative system. The distillation aims to encourage the single model to learn from and imitate the collaboration trace of the multi-LM system: at inference time, only the distilled model is employed for generation, improving the efficiency by cutting the cost from multiple LMs back to one while maintaining the benefits of collaboration.

\begin{figure*}[t]
    \centering
    \includegraphics[width=1.0\linewidth]{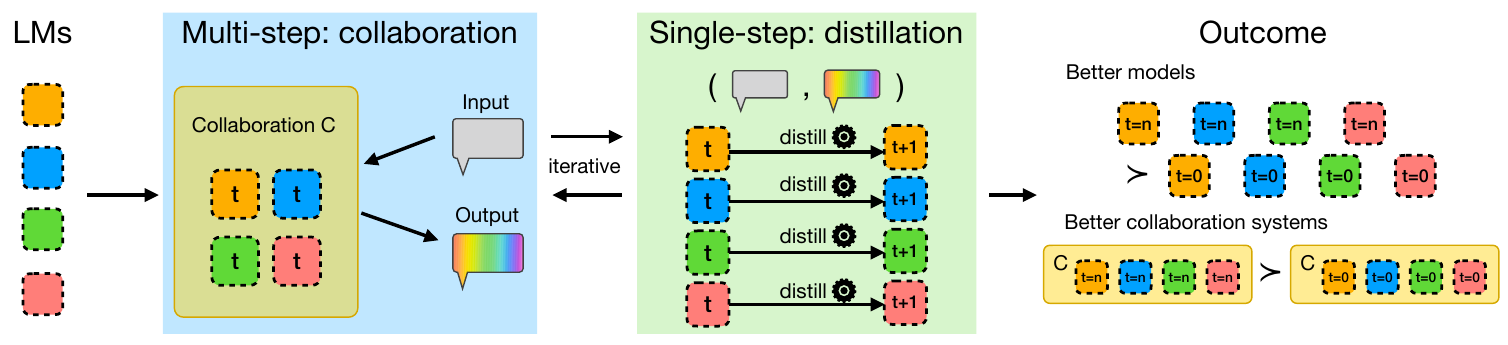}
    \caption{We propose the single-multi evolution loop: In the \emph{multi-step}, multiple language models collaborate via model collaboration algorithm $\mathcal{C}$ to generate better responses; In the \emph{single-step}, we employ knowledge distillation, where each individual LM is the student and the model collaboration system is the teacher. By alternating and iteratively executing the multi-step and single-step, multiple LLMs collaboratively evolve for better models and better model collaboration systems.}
    \label{fig:overview}
\end{figure*}

Furthermore, we propose the \emph{single-multi evolution loop} where multiple LMs evolve through iteratively executing this collaboration-distillation duo: models first \emph{collaborate}, each of the LMs then independently \emph{distill} from the collaborative outputs, and these post-distillation, enhanced, and more synergistic LMs then \emph{collaborate} again, iteratively. In this way, language models evolve by \emph{interacting and learning from a system of other models}: this would benefit both the capacity of the individual LMs and the collaboration system built on top of these evolving language models.

We conduct extensive experiments with 3 model pool settings, 7 collaboration strategies, 3 distillation methods, and evaluate on 15 tasks spanning QA, reasoning, factuality, safety, instruction following, and more. Extensive experiments demonstrate that: 1) distilling from model collaboration improves individual LMs by 8.0\% on average, and by using them in place of the multi-LM systems we improve efficiency while preserving collaborative gains; 2) the multi-LM collaboration systems also benefit from having the stronger and evolved LMs as components, improving over initial unevolved systems by 14.9\% across tasks. Further analysis reveals that the single-multi evolution loop outperforms various existing evolutionary strategies by 7.7\% on average, helps solve an average of 66.5\% of problems in QA and reasoning where the initial models/systems struggle, and works with diverse models, collaboration methods, and distillation algorithms. Together, these efforts put forward a new paradigm of evolutionary AI systems: models self-improve and evolve by collaborating and interacting with an environment of other models.

\section{Methodology}

\subsection{Overview}

\begin{algorithm}[t]
\caption{Single--Multi Evolution Loop}
\label{alg:single-multi-evolution}
\begin{algorithmic}[1]
\REQUIRE 
Initial model pool $\mathcal{M}^{(0)} = \{\mathbf{m}_1^{(0)}, \cdots, \mathbf{m}_n^{(0)}\}$; \\
Collaboration strategy $\mathcal{C}$ (e.g., Multi-agent Debate); \\
Distillation method $\mathcal{D}$ (e.g., Supervised KD); \\
Instruction dataset $\mathcal{X}$; 
Number of iterations $k$
\ENSURE 
Evolved single model pool $\mathcal{M}^{(k)}$ and model collaboration system $\mathcal{C}(\mathcal{M}^{(k)})$

\FOR{$t = 0$ to $k-1$}

    \STATE \textbf{Multi-step: Collaboration}
    \STATE Initialize empty dataset $\mathcal{D}^{(t)}$
    \FORALL{$\mathbf{x} \in \mathcal{X}$}
        \STATE $\mathbf{y} \leftarrow \mathcal{C}(\mathbf{x} \mid \mathcal{M}^{(t)})$
        \STATE $\mathcal{D}^{(t)} \leftarrow \mathcal{D}^{(t)} \cup \{(\mathbf{x}, \mathbf{y})\}$
    \ENDFOR

    \STATE \textbf{Single-step: Distillation}
    \FORALL{$\mathbf{m}_i^{(t)} \in \mathcal{M}^{(t)}$}
        \STATE $\mathbf{m}_i^{(t+1)} \leftarrow \mathcal{D}\big(\mathbf{m}_i^{(t)} \mid \mathcal{D}^{(t)}\big)$
    \ENDFOR

    \STATE $\mathcal{M}^{(t+1)} \leftarrow \{\mathbf{m}_1^{(t+1)}, \cdots, \mathbf{m}_n^{(t+1)}\}$

\ENDFOR

\textbf{Return} $\mathcal{M}^{(k)}, \mathcal{C}(\mathcal{M}^{(k)})$
\end{algorithmic}
\end{algorithm}

Given language models $\mathcal{M}=\{\mathbf{m}_1, \cdots, \mathbf{m}_n\}$ and natural language instructions $\mathbf{x} \in \mathcal{X}$, we propose to let models $\mathcal{M}$ collaborate (the \emph{multi-step}), distill the collaboration outcome into individual models $\mathbf{m} \in \mathcal{M}$ for efficiency (the \emph{single-step}), and employ the post-distillation improved LMs for collaboration again. By alternating between the multi-step and the single-step, language models collectively evolve by interacting with an environment of other models.

In the \emph{multi-step}, given instruction $\mathbf{x} \in \mathcal{X}$, we employ model collaboration strategy $\mathcal{C}$ (e.g., multiagent debate \citep{du2023improving} or routing \citep{ongroutellm}, Section \ref{subsec:collaboration}) to generate a response $\mathcal{C}(\mathbf{x} \mid \mathcal{M}^{(t)})$, where $(t)$ denotes the iteration time stamp. Marginalizing over $\mathbf{x} \in \mathcal{X}$, we obtain a supervised fine-tuning dataset with outputs generated by the model collaboration system:

\begin{align*}
    \mathcal{D}^{(t)} = \{(\mathbf{x}, \mathcal{C}(\mathbf{x} \mid \mathcal{M}^{(t)}))\}_{\mathbf{x} \in \mathcal{X}}
\end{align*}

In the \emph{single-step}, we distill the collaborative outputs into each individual model (e.g., supervised distillation \citep{kim2016sequence} or on-policy distillation \citep{lin2020autoregressive, agarwal2024policy}), thus learn from the collaboration patterns while only incurring the cost of a single model when used for inference. Specifically:

\begin{align*}
    \mathbf{m}_i^{(t+1)} = \mathrm{distill}(\mathbf{m}_i^{(t)} \mid \mathcal{D}^{(t)})
\end{align*}
This yields a pool of improved language models $\mathcal{M}^{(t+1)}=\{\mathbf{m}_1^{(t+1)}, \cdots, \mathbf{m}_n^{(t+1)}\}$ after interacting and learning with the collaborative system of multiple models.

The \emph{single-multi evolution loop} is then an iterative process of alternating between the single-step and the multi-step:

\begin{align*}
    \text{LMs} & \ \mathcal{M}^{(0)} \xrightarrow[]{\text{multi}} \text{system} \ \mathcal{C}(\mathcal{M}^{(0)}) \xrightarrow[]{\text{single}} \text{LMs} \ \mathcal{M}^{(1)} \\ & \xrightarrow[]{\text{multi}} \cdots \xrightarrow[]{\text{single}} \text{LMs} \ \mathcal{M}^{(k)} \xrightarrow[]{\text{multi}} \text{system} \ \mathcal{C}(\mathcal{M}^{(k)})
\end{align*}
where $k$ denotes the number of iterations. In this way, models in $\mathcal{M}$ collectively evolve and self-improve by interacting and collaborating with other language models. The objective of the \emph{single-multi evolution loop} is twofold:

\begin{align*}
    \mathcal{M}^{(k)} \succ \mathcal{M}^{(0)} \ \ \& \ \ \mathcal{C}(\mathcal{M}^{(k)}) \succ \mathcal{C}(\mathcal{M}^{(0)})
\end{align*}
that both the individual models and the collaboration system consisting of these models should improve compared to the initial models and system, where $\succ$ indicates better performance. We provide an overview of the single-multi evolution loop in Figure \ref{fig:overview} and Algorithm \ref{alg:single-multi-evolution}.

\subsection{Multi-step: collaboration strategies $\mathcal{C}$}
\label{subsec:collaboration}
In the multi-step, multiple LMs $\mathcal{M}$ should collaborate and complement each other to generate a better response for instruction $\mathbf{x} \in \mathcal{X}$. We seek to cover the wide range of literature on model collaboration \citep{feng2025one} by employing one model collaboration method per collaboration level, while experimenting with more in Section \ref{sec:analysis}.

\emph{API-level collaboration} features the routing and selection of different LMs for different user queries \citep{ongroutellm, frick2025prompt, fenggraphrouter,zheng2025citer}. At training time, we evaluate which candidate model would generate the best response to $\mathbf{x}$: $j = \arg \max_{i=1}^n \mathrm{score}(\mathbf{m}_i(\mathbf{x}))$, then train a router $f$ on $f(\mathbf{x} \mid \mathcal{M}) \rightarrow \mathbf{m}_j$. At inference-time, we employ the model $f(\mathbf{x}' \mid \mathcal{M})$ to generate responses for $\mathbf{x}'$. We instantiate $f$ as a causal language model \citep{ongroutellm}.

\emph{Text-level collaboration} features the discussion and debate among multiple LMs to give feedback and refine responses \citep{guo2024large, zhao2025language, dang2025multi}. We specifically employ multi-agent debate \citep{du2023improving} where models refine responses based on others' responses with the following prompt:

\begin{center}
\begin{tcolorbox}[colback=mygray!5!white, colframe=myblue!75!black, width=1.0\linewidth, title=Prompt 1: collaboration with multi-agent debate, coltitle=white, colbacktitle=myblue]
    \small You are part of a team of AI assistants collaborating to answer the user's question. Each assistant provides their own answer: use their answers to refine and improve your own answer.\\
    Question: (question)\\
    Your previous answer: (previous answer)\\
    Other assistants' answers: (list of answers, one per row)\\
    Please provide a refined answer to the question.
\end{tcolorbox}
\end{center}
We run this for a few iterations and employ the best model on a held-out set to summarize all models' responses at the last iteration for a final response.

\emph{Logit-level collaboration} features arithmetic with the next-token probability distributions across multiple LMs for collaborative decoding \citep{liutuning, shen2024learning}. We specifically employ logit fusion: for model $\mathbf{m}_i$'s next-token distribution $\mathbf{p}_i$, we aggregate across models $\mathbf{p} = \frac{1}{n} \sum_{i=1}^n \mathbf{p}_i$ and then decode text from $\mathbf{p}$. This requires models in $\mathcal{M}$ to share mostly similar vocabulary/tokenization (e.g., LoRA adapters \citep{hulora} finetuned from a common base model).

\emph{Weight-level collaboration} features the merging and searching in the model parameter space across multiple LMs \citep{huanglorahub, yadavsurvey, feng2025model}. We specifically employ dare-ties model merging \citep{yadav2023ties, yu2024language} due to their strong numerical properties. $\mathcal{C}(\mathbf{x} \mid \mathcal{M})$ in this case is then the response generated by the merged model. This requires models in $\mathcal{M}$ to share mostly the same architecture (e.g., LoRA adapters finetuned from a common base model).

With the four representative model collaboration strategies as $\mathcal{C}$, we aim to investigate whether the \emph{single-model evolution loop} is a broadly applicable idea and whether certain modes of collaboration would be especially helpful.

\subsection{Single-step: distillation strategies}
\label{subsec:distillation}
In the single-step, we independently employ each model in $\mathcal{M}$ as student and the collaboration system $\mathcal{C}(\mathbf{x} \mid \mathcal{M})$ as teacher for knowledge distillation (KD), so models could interact with other models and learn from the collaboration outcome to collectively evolve and self-improve. We consider the following distillation strategies:

\emph{Supervised KD}: model $\mathbf{m}$ is supervised fine-tuned on the generated texts of the collaboration system $\{(\mathbf{x}, \mathcal{C}(\mathbf{x} \mid \mathcal{M}^{(t)}))\}_{\mathbf{x} \in \mathcal{X}}$. This is most flexible and efficient, without making assumptions about tokenization.

\emph{Multi-student KD} \citep{lei2025learning}: two findings motivate an adjusted version of supervised KD: 1) the skill gap between $\mathbf{m}$ and $\mathcal{C}(\mathbf{x} \mid \mathcal{M})$ is sometimes large for effective distillation, especially in early iterations of the single-multi evolution loop; 2) since all models employ the same teacher/data for distillation, the variation/diversity across models might be reduced. As such, we formulate multi-student KD, where the models are supervised fine-tuned on an $\alpha : \beta : \gamma$ mixture of collaboration outputs, outputs of the strongest student (with a smaller student-student capability gap than teacher-student), and outputs of itself (to preserve its specialization), respectively $\{(\mathbf{x}, \mathcal{C}(\mathbf{x} \mid \mathcal{M}^{(t)}))\}_{\mathbf{x} \in \mathcal{X}}$, $\{(\mathbf{x}, \mathbf{m}_{\textit{best}}(\mathbf{x}))\}_{\mathbf{x} \in \mathcal{X}}$, and $\{(\mathbf{x}, \mathbf{m}(\mathbf{x})\}_{\mathbf{x} \in \mathcal{X}}$.

\emph{Logit-based KD}: we additionally employ logit-based distillation (minimizing the KL divergence between student/teacher logit distributions, in addition to cross-entropy loss of token prediction), specifically on-policy distillation \citep{agarwal2024policy, patiño2025_unlocking_on_policy_distillation_for_any_model_family} (token from student but logit from teacher). These methods are bounded by tokenization assumptions between the teacher and the student for logit calculations and may not be applicable to every collaboration strategy $\mathcal{C}$ as the teacher, potentially extensible by reconciling tokenization differences.

\begin{table*}[t]
\centering
\scriptsize
\setlength{\tabcolsep}{1pt}
\renewcommand{\arraystretch}{1.4}
\vspace*{5pt}
\caption{Performance of the single-multi evolution loop with different model pools, collaboration methods, and distillation settings. We highlight improvements in \colorbox{orange!15}{orange} ($\mathrm{single}_{t=n} > \mathrm{single}_{t=0}$, $\mathrm{multi}_{t=n} > \mathrm{multi}_{t=0}$). Best in \textbf{bold} and second-best in \underline{underline}. Results show that the single-multi evolution loop is a broadly effective idea, compatible with diverse models and collaboration strategies, bringing performance gains to both individual models and collaboration systems in 91.9\% settings by 8.0\% and 14.9\% on average.}
\vspace*{-5pt}
\label{tab:big}
\resizebox{1\linewidth}{!}{ 

\begin{tabular}{lccccc|cccccc|ccc|ccc}\toprule[1.5pt]
& & &\multicolumn{15}{c}{pool 1: specialized LLMs} \\\cmidrule{4-18}
& & &\multicolumn{3}{c}{general QA} &\multicolumn{6}{c}{reasoning} &\multicolumn{3}{c}{knowledge} &safety &science & inst. \\\cmidrule{4-18}
& & &agieval &arc &mmlu-r &bbh &gsm8k &math &gpqa-dia &tablemwp &theoremqa &wikidyk &popqa &blend &truthful &sciriff &hinterest \\\midrule
\multicolumn{3}{c}{single, t=0} &51.10 {\tiny (0.72)} &81.66 {\tiny (3.94)} &65.57 {\tiny (2.21)} &38.53 {\tiny (7.13)} &82.40 {\tiny (2.36)} &77.41 {\tiny (0.79)} &31.99 {\tiny (1.54)} &61.87 {\tiny (2.01)} &27.92 {\tiny (1.91)} &4.43 {\tiny (0.35)} &15.79 {\tiny (1.28)} &72.23 {\tiny (1.47)} &56.83 {\tiny (5.49)} &53.48 {\tiny (6.20)} &1.80 {\tiny (0.68)} \\
\multirow{4}{*}{multi, t=0} &\multicolumn{2}{c}{api: trained router} &51.04 &81.06 &65.10 &39.30 &79.64 &74.93 &27.27 &63.20 &34.00 &4.87 &16.27 &70.70 &54.46 &52.52 &2.24 \\
&\multicolumn{2}{c}{text: multiagent debate} &52.77 &68.94 &55.50 &45.90 &76.50 &81.28 &27.27 &42.50 &30.25 &3.33 &14.66 &67.10 &45.06 &54.68 &2.98 \\
&\multicolumn{2}{c}{logit: logit fusion} &43.43 &83.79 &73.70 &13.10 &55.70 &68.05 &34.34 &43.20 &21.50 &5.97 &7.43 &61.60 &51.70 &47.48 &-3.14 \\
&\multicolumn{2}{c}{weight: dare ties} &52.16 &82.17 &69.90 &40.90 &83.90 &71.40 &29.29 &60.80 &25.00 &5.08 &17.77 &73.00 &51.28 &\underline{64.75} &0.03 \\ \midrule[0.75pt]
\multirow{8}{*}{supervised-KD} &\multirow{2}{*}{api} &single, t=n &\cellcolor{orange!15}\underline{62.31} {\tiny (0.40)} &\cellcolor{orange!15}86.58 {\tiny (0.63)} &\cellcolor{orange!15}71.10 {\tiny (0.78)} &\cellcolor{orange!15}40.40 {\tiny (0.36)} &\cellcolor{orange!15}90.37 {\tiny (0.76)} &\cellcolor{orange!15}81.76 {\tiny (0.22)} &30.98 {\tiny (1.17)} &\cellcolor{orange!15}72.27 {\tiny (1.67)} &\cellcolor{orange!15}33.17 {\tiny (1.63)} &\cellcolor{orange!15}5.60 {\tiny (0.24)} &\cellcolor{orange!15}18.14 {\tiny (0.41)} &\cellcolor{orange!15}78.83 {\tiny (0.12)} &\cellcolor{orange!15}67.37 {\tiny (0.50)} &\cellcolor{orange!15}56.59 {\tiny (1.10)} &\cellcolor{orange!15}15.27 {\tiny (0.50)} \\
& &multi, t=n &\cellcolor{orange!15}62.28 &\cellcolor{orange!15}86.52 &\cellcolor{orange!15}70.60 &\cellcolor{orange!15}40.90 &\cellcolor{orange!15}89.90 &\cellcolor{orange!15}\underline{82.22} &\cellcolor{orange!15}33.33 &\cellcolor{orange!15}\underline{73.60} &\cellcolor{orange!15}36.75 &\cellcolor{orange!15}5.27 &\cellcolor{orange!15}\textbf{20.92} &\cellcolor{orange!15}79.00 &\cellcolor{orange!15}\underline{68.88} &\cellcolor{orange!15}59.71 &\cellcolor{orange!15}15.14 \\
&\multirow{2}{*}{text} &single, t=n &\cellcolor{orange!15}55.42 {\tiny (0.53)} &\cellcolor{orange!15}84.78 {\tiny (0.69)} &\cellcolor{orange!15}68.20 {\tiny (0.70)} &\cellcolor{orange!15}42.20 {\tiny (0.26)} &\cellcolor{orange!15}90.70 {\tiny (1.00)} &\cellcolor{orange!15}81.97 {\tiny (0.63)} &30.30 {\tiny (1.01)} &\cellcolor{orange!15}72.53 {\tiny (0.46)} &\cellcolor{orange!15}33.58 {\tiny (1.70)} &\cellcolor{orange!15}5.60 {\tiny (0.22)} &\cellcolor{orange!15}18.10 {\tiny (0.34)} &\cellcolor{orange!15}78.97 {\tiny (0.61)} &\cellcolor{orange!15}67.04 {\tiny (0.80)} &\cellcolor{orange!15}57.31 {\tiny (0.42)} &\cellcolor{orange!15}\underline{15.28} {\tiny (0.38)} \\
& &multi, t=n &\cellcolor{orange!15}57.96 &\cellcolor{orange!15}72.27 &\cellcolor{orange!15}60.50 &\cellcolor{orange!15}51.60 &\cellcolor{orange!15}85.00 &\cellcolor{orange!15}82.11 &25.25 &\cellcolor{orange!15}69.60 &\cellcolor{orange!15}\underline{39.06} &\cellcolor{orange!15}5.92 &\cellcolor{orange!15}16.70 &\cellcolor{orange!15}74.80 &\cellcolor{orange!15}68.07 &\cellcolor{orange!15}58.27 &\cellcolor{orange!15}9.42 \\
&\multirow{2}{*}{logit} &single, t=n &\cellcolor{orange!15}60.64 {\tiny (1.36)} &\cellcolor{orange!15}85.61 {\tiny (0.56)} &\cellcolor{orange!15}68.40 {\tiny (0.40)} &\cellcolor{orange!15}40.60 {\tiny (0.66)} &68.13 {\tiny (3.71)} &71.61 {\tiny (7.13)} &\cellcolor{orange!15}\underline{38.62} {\tiny (3.25)} &\cellcolor{orange!15}71.20 {\tiny (1.22)} &24.58 {\tiny (3.17)} &\cellcolor{orange!15}5.30 {\tiny (0.14)} &\cellcolor{orange!15}17.88 {\tiny (0.21)} &\cellcolor{orange!15}79.07 {\tiny (0.35)} &\cellcolor{orange!15}67.86 {\tiny (1.04)} &\cellcolor{orange!15}56.83 {\tiny (0.72)} &-0.43 {\tiny (0.21)} \\
& &multi, t=n &35.99 &\cellcolor{orange!15}86.18 &27.10 &\cellcolor{orange!15}43.19 &\cellcolor{orange!15}56.50 &\cellcolor{orange!15}71.73 &\cellcolor{orange!15}\textbf{40.94} &\cellcolor{orange!15}48.80 &21.50 &\cellcolor{orange!15}\textbf{8.10} &\cellcolor{orange!15}18.55 &\cellcolor{orange!15}76.20 &\cellcolor{orange!15}66.29 &\cellcolor{orange!15}58.27 &\cellcolor{orange!15}0.74 \\
&\multirow{2}{*}{weight} &single, t=n &44.87 {\tiny (0.95)} &\cellcolor{orange!15}85.64 {\tiny (0.39)} &\cellcolor{orange!15}69.63 {\tiny (0.57)} &\cellcolor{orange!15}41.13 {\tiny (1.36)} &\cellcolor{orange!15}90.70 {\tiny (0.72)} &\cellcolor{orange!15}81.90 {\tiny (0.55)} &\cellcolor{orange!15}32.66 {\tiny (2.10)} &\cellcolor{orange!15}71.73 {\tiny (1.67)} &\cellcolor{orange!15}33.00 {\tiny (1.09)} &\cellcolor{orange!15}5.25 {\tiny (0.22)} &\cellcolor{orange!15}18.23 {\tiny (0.13)} &\cellcolor{orange!15}78.80 {\tiny (0.56)} &\cellcolor{orange!15}67.21 {\tiny (0.47)} &\cellcolor{orange!15}57.07 {\tiny (1.66)} &\cellcolor{orange!15}15.15 {\tiny (0.68)} \\
& &multi, t=n &\cellcolor{orange!15}57.18 &\cellcolor{orange!15}83.19 &\cellcolor{orange!15}\underline{73.90} &\cellcolor{orange!15}\underline{56.14} &82.90 &\cellcolor{orange!15}77.30 &25.25 &\cellcolor{orange!15}\textbf{78.77} &\cellcolor{orange!15}27.75 &\cellcolor{orange!15}\underline{7.61} &\cellcolor{orange!15}\underline{19.78} &\cellcolor{orange!15}74.85 &\cellcolor{orange!15}\textbf{69.60} &\cellcolor{orange!15}\textbf{70.60} &\cellcolor{orange!15}12.24 \\ \midrule[0.75pt]
\multirow{8}{*}{multi-KD} &\multirow{2}{*}{api} &single, t=n &\cellcolor{orange!15}61.19 {\tiny (1.10)} &\cellcolor{orange!15}86.29 {\tiny (0.20)} &\cellcolor{orange!15}70.8 {\tiny (0.82)} &\cellcolor{orange!15}40.23 {\tiny (0.49)} &\cellcolor{orange!15}89.77 {\tiny (0.49)} &\cellcolor{orange!15}81.97 {\tiny (0.63)} &\cellcolor{orange!15}34.01 {\tiny (1.17)} &\cellcolor{orange!15}70.67 {\tiny (0.46)} &\cellcolor{orange!15}33.92 {\tiny (1.26)} &\cellcolor{orange!15}5.36 {\tiny (0.19)} &\cellcolor{orange!15}16.08 {\tiny (2.74)} &\cellcolor{orange!15}78.70 {\tiny (0.26)} &\cellcolor{orange!15}66.56 {\tiny (0.57)} &\cellcolor{orange!15}56.35 {\tiny (0.42)} &\cellcolor{orange!15}14.85 {\tiny (0.44)} \\
& &multi, t=n &\cellcolor{orange!15}\textbf{66.77} &\cellcolor{orange!15}87.03 &\cellcolor{orange!15}71.30 &\cellcolor{orange!15}\textbf{58.20} &\cellcolor{orange!15}\textbf{91.20} &\cellcolor{orange!15}77.09 &\cellcolor{orange!15}32.32 &\cellcolor{orange!15}72.80 &\cellcolor{orange!15}34.75 &\cellcolor{orange!15}5.79 &15.82 &\cellcolor{orange!15}\textbf{79.60} &\cellcolor{orange!15}68.23 &\cellcolor{orange!15}57.55 &\cellcolor{orange!15}15.53 \\
&\multirow{2}{*}{text} &single, t=n &\cellcolor{orange!15}59.17 {\tiny (0.79)} &\cellcolor{orange!15}86.18 {\tiny (0.96)} &\cellcolor{orange!15}69.03 {\tiny (0.81)} &\cellcolor{orange!15}40.83 {\tiny (1.43)} &\cellcolor{orange!15}\underline{90.77} {\tiny (0.21)} &\cellcolor{orange!15}82.15 {\tiny (0.59)} &\cellcolor{orange!15}34.68 {\tiny (4.77)} &\cellcolor{orange!15}73.07 {\tiny (1.22)} &\cellcolor{orange!15}34.42 {\tiny (1.63)} &\cellcolor{orange!15}5.80 {\tiny (0.24)} &\cellcolor{orange!15}18.30 {\tiny (0.06)} &\cellcolor{orange!15}\underline{79.10} {\tiny (0.50)} &\cellcolor{orange!15}67.59 {\tiny (0.58)} &\cellcolor{orange!15}57.55 {\tiny (0.72)} &\cellcolor{orange!15}15.05 {\tiny (0.40)} \\
& &multi, t=n &\cellcolor{orange!15}55.19 &\cellcolor{orange!15}73.81 &\cellcolor{orange!15}\textbf{79.20} &\cellcolor{orange!15}50.10 &\cellcolor{orange!15}85.20 &\cellcolor{orange!15}\textbf{83.37} &\cellcolor{orange!15}31.31 &\cellcolor{orange!15}71.20 &\cellcolor{orange!15}\textbf{41.40} &\cellcolor{orange!15}6.01 &\cellcolor{orange!15}16.14 &\cellcolor{orange!15}71.60 &\cellcolor{orange!15}66.61 &\cellcolor{orange!15}58.27 &\cellcolor{orange!15}10.01 \\
&\multirow{2}{*}{logit} &single, t=n &\cellcolor{orange!15}53.84 {\tiny (1.48)} &\cellcolor{orange!15}\underline{87.34} {\tiny (0.27)} &\cellcolor{orange!15}69.97 {\tiny (0.25)} &\cellcolor{orange!15}40.50 {\tiny (0.44)} &67.07 {\tiny (2.65)} &61.96 {\tiny (7.56)} &26.26 {\tiny (6.14)} &\cellcolor{orange!15}71.20 {\tiny (1.60)} &26.00 {\tiny (0.90)} &\cellcolor{orange!15}5.64 {\tiny (0.23)} &\cellcolor{orange!15}18.70 {\tiny (0.25)} &\cellcolor{orange!15}78.47 {\tiny (0.45)} &\cellcolor{orange!15}67.21 {\tiny (1.04)} &\cellcolor{orange!15}56.59 {\tiny (1.50)} &\cellcolor{orange!15}4.47 {\tiny (1.69)} \\
& &multi, t=n &43.08 &\cellcolor{orange!15}84.73 &45.24 &\cellcolor{orange!15}17.10 &\cellcolor{orange!15}60.50 &46.23 &\cellcolor{orange!15}37.37 &\cellcolor{orange!15}53.60 &21.50 &1.20 &6.04 &\cellcolor{orange!15}77.00 &\cellcolor{orange!15}65.96 &\cellcolor{orange!15}57.55 &\cellcolor{orange!15}8.65 \\
&\multirow{2}{*}{weight} &single, t=n &\cellcolor{orange!15}60.73 {\tiny (0.17)} &\cellcolor{orange!15}86.63 {\tiny (0.50)} &\cellcolor{orange!15}71.67 {\tiny (0.55)} &\cellcolor{orange!15}39.93 {\tiny (0.68)} &\cellcolor{orange!15}89.67 {\tiny (0.81)} &\cellcolor{orange!15}82.04 {\tiny (0.79)} &29.63 {\tiny (4.98)} &\cellcolor{orange!15}72.53 {\tiny (1.67)} &\cellcolor{orange!15}34.33 {\tiny (0.88)} &\cellcolor{orange!15}5.62 {\tiny (0.30)} &\cellcolor{orange!15}18.82 {\tiny (0.45)} &\cellcolor{orange!15}78.40 {\tiny (0.30)} &\cellcolor{orange!15}67.31 {\tiny (0.61)} &\cellcolor{orange!15}56.35 {\tiny (0.42)} &\cellcolor{orange!15}\textbf{15.31} {\tiny (0.29)} \\
& &multi, t=n &\cellcolor{orange!15}56.57 &\cellcolor{orange!15}\textbf{87.46} &\cellcolor{orange!15}71.30 &40.40 &\cellcolor{orange!15}\textbf{91.20} &\cellcolor{orange!15}79.29 &\cellcolor{orange!15}34.34 &\cellcolor{orange!15}70.40 &\cellcolor{orange!15}33.50 &5.08 &17.67 &\cellcolor{orange!15}78.80 &\cellcolor{orange!15}68.40 &57.55 &\cellcolor{orange!15}18.73 \\ \midrule[0.75pt]
& & &\multicolumn{15}{c}{pool 2: general-purpose LLMs} \\ \midrule[0.75pt]
\multicolumn{3}{c}{single, t=0} &38.87 {\tiny (6.28)} &61.77 {\tiny (7.69)} &50.87 {\tiny (8.47)} &32.40 {\tiny (9.72)} &83.47 {\tiny (6.67)} &72.97 {\tiny (10.21)} &25.93 {\tiny (3.55)} &60.80 {\tiny (10.82)} &27.08 {\tiny (5.20)} &4.73 {\tiny (2.43)} &16.47 {\tiny (9.04)} &57.37 {\tiny (3.80)} &50.08 {\tiny (7.97)} &48.68 {\tiny (8.90)} &2.09 {\tiny (1.86)} \\
\multirow{2}{*}{multi, t=0} &\multicolumn{2}{c}{api: trained router} &38.33 &62.65 &52.20 &54.20 &72.22 &74.54 &25.25 &49.60 &30.25 &7.91 &17.41 {\tiny (9.38)} &65.70 &49.81 &35.70 &0.26 \\
&\multicolumn{2}{c}{text: multiagent debate} &54.67 &63.65 &51.60 &60.40 &83.30 &74.58 &22.22 &68.00 &30.00 &5.53 &19.78 &68.00 &57.37 &55.40 &6.74 \\ \midrule[0.75pt]
\multirow{4}{*}{supervised-KD} &\multirow{2}{*}{api} &single, t=n &\cellcolor{orange!15}40.66 {\tiny (5.84)} &\cellcolor{orange!15}64.56 {\tiny (9.88)} &\cellcolor{orange!15}53.57 {\tiny (4.41)} &\cellcolor{orange!15}35.50 {\tiny (5.30)} &\cellcolor{orange!15}83.57 {\tiny (6.33)} &\cellcolor{orange!15}77.01 {\tiny (9.66)} &\cellcolor{orange!15}30.30 {\tiny (2.67)} &\cellcolor{orange!15}62.40 {\tiny (11.00)} &\cellcolor{orange!15}29.08 {\tiny (6.42)} &\cellcolor{orange!15}7.08 {\tiny (2.62)} &\cellcolor{orange!15}19.96 {\tiny (9.66)} &\cellcolor{orange!15}59.20 {\tiny (3.98)} &\cellcolor{orange!15}52.78 {\tiny (8.43)} &\cellcolor{orange!15}53.00 {\tiny (5.45)} &\cellcolor{orange!15}2.80 {\tiny (1.88)} \\
& &multi, t=n &\cellcolor{orange!15}41.75 &\cellcolor{orange!15}\textbf{70.65} &\cellcolor{orange!15}\textbf{55.20} &\cellcolor{orange!15}\textbf{67.10} &\cellcolor{orange!15}79.80 &\cellcolor{orange!15}\underline{78.49} &\cellcolor{orange!15}\textbf{37.23} &\cellcolor{orange!15}67.60 &\cellcolor{orange!15}\textbf{39.90} &\cellcolor{orange!15}\textbf{12.75} &\cellcolor{orange!15}\underline{21.33} &\cellcolor{orange!15}71.40 &\cellcolor{orange!15}57.00 &\cellcolor{orange!15}\textbf{68.60} &\cellcolor{orange!15}1.37 \\
&\multirow{2}{*}{text} &single, t=n &\cellcolor{orange!15}40.54 {\tiny (3.83)} &\cellcolor{orange!15}64.48 {\tiny (4.59)} &\cellcolor{orange!15}52.50 {\tiny (3.26)} &\cellcolor{orange!15}34.63 {\tiny (9.56)} &83.40 {\tiny (6.08)} &\cellcolor{orange!15}76.43 {\tiny (10.00)} &\cellcolor{orange!15}27.95 {\tiny (4.55)} &\cellcolor{orange!15}63.47 {\tiny (10.41)} &\cellcolor{orange!15}28.58 {\tiny (4.58)} &\cellcolor{orange!15}5.27 {\tiny (3.11)} &\cellcolor{orange!15}19.08 {\tiny (9.64)} &\cellcolor{orange!15}58.80 {\tiny (3.74)} &\cellcolor{orange!15}51.32 {\tiny (8.02)} &\cellcolor{orange!15}53.24 {\tiny (8.72)} &\cellcolor{orange!15}2.71 {\tiny (1.88)} \\
& &multi, t=n &\cellcolor{orange!15}\textbf{57.53} &\cellcolor{orange!15}65.10 &\cellcolor{orange!15}53.22 &\cellcolor{orange!15}61.70 &\cellcolor{orange!15}\underline{85.80} &\cellcolor{orange!15}77.30 &\cellcolor{orange!15}34.34 &\cellcolor{orange!15}\underline{72.00} &\cellcolor{orange!15}\underline{32.75} &\cellcolor{orange!15}5.65 &\cellcolor{orange!15}20.06 &\cellcolor{orange!15}\underline{71.70} &\cellcolor{orange!15}\underline{59.16} &\cellcolor{orange!15}62.59 &\cellcolor{orange!15}\underline{7.05} \\ \midrule[0.75pt]
\multirow{4}{*}{multi-KD} &\multirow{2}{*}{api} &single, t=n &\cellcolor{orange!15}45.70 {\tiny (6.86)} &\cellcolor{orange!15}65.48 {\tiny (7.49)} &\cellcolor{orange!15}53.30 {\tiny (8.74)} &\cellcolor{orange!15}34.30 {\tiny (6.45)} &\cellcolor{orange!15}83.67 {\tiny (6.05)} &\cellcolor{orange!15}76.03 {\tiny (9.66)} &\cellcolor{orange!15}28.96 {\tiny (2.33)} &\cellcolor{orange!15}62.93 {\tiny (10.13)} &\cellcolor{orange!15}30.58 {\tiny (3.01)} &\cellcolor{orange!15}5.22 {\tiny (2.50)} &\cellcolor{orange!15}19.58 {\tiny (8.92)} &\cellcolor{orange!15}58.77 {\tiny (3.66)} &\cellcolor{orange!15}51.38 {\tiny (7.31)} &\cellcolor{orange!15}52.52 {\tiny (8.11)} &\cellcolor{orange!15}2.66 {\tiny (1.91)} \\
& &multi, t=n &\cellcolor{orange!15}38.49 &\cellcolor{orange!15}66.20 &\cellcolor{orange!15}\underline{53.60} &\cellcolor{orange!15}\underline{65.30} &\cellcolor{orange!15}78.30 &\cellcolor{orange!15}75.59 &\cellcolor{orange!15}27.07 &\cellcolor{orange!15}\textbf{72.40} &\cellcolor{orange!15}31.32 &\cellcolor{orange!15}\underline{8.55} &\cellcolor{orange!15}\textbf{22.15} &\cellcolor{orange!15}71.20 &\cellcolor{orange!15}57.74 &\cellcolor{orange!15}\underline{65.90} &\cellcolor{orange!15}0.53 \\
&\multirow{2}{*}{text} &single, t=n &\cellcolor{orange!15}45.04 {\tiny (6.36)} &\cellcolor{orange!15}65.10 {\tiny (5.51)} &\cellcolor{orange!15}53.10 {\tiny (7.27)} &\cellcolor{orange!15}35.90 {\tiny (3.28)} &\cellcolor{orange!15}84.50 {\tiny (5.21)} &\cellcolor{orange!15}75.84 {\tiny (10.09)} &\cellcolor{orange!15}29.63 {\tiny (0.58)} &\cellcolor{orange!15}61.33 {\tiny (10.07)} &\cellcolor{orange!15}29.17 {\tiny (4.13)} &\cellcolor{orange!15}5.23 {\tiny (2.37)} &\cellcolor{orange!15}19.20 {\tiny (9.55)} &\cellcolor{orange!15}58.97 {\tiny (3.95)} &\cellcolor{orange!15}51.22 {\tiny (7.49)} &\cellcolor{orange!15}53.24 {\tiny (6.86)} &\cellcolor{orange!15}2.50 {\tiny (1.88)} \\
& &multi, t=n &\cellcolor{orange!15}\underline{57.27} &\cellcolor{orange!15}\underline{67.06} &\cellcolor{orange!15}52.50 &\cellcolor{orange!15}\underline{65.30} &\cellcolor{orange!15}\textbf{86.30} &\cellcolor{orange!15}\textbf{80.13} &\cellcolor{orange!15}\underline{36.36} &\cellcolor{orange!15}\underline{72.00} &\cellcolor{orange!15}\underline{32.75} &\cellcolor{orange!15}6.17 &\cellcolor{orange!15}20.35 &\cellcolor{orange!15}\textbf{72.50} &\cellcolor{orange!15}\textbf{61.10} &\cellcolor{orange!15}62.59 &\cellcolor{orange!15}\textbf{7.23} \\
\bottomrule[1.5pt]
\end{tabular}
\vspace*{-10pt}
}
\end{table*}

\section{Experiment Settings}

\paragraph{Models and Implementation} We employ two model pools for collaboration and distillation: three specialized language models based on \textsc{Qwen2.5-7B-Instruct}, separately finetuned \citep{jiang2025sparta} on three domains of SFT data in Tulu-v3 \citep{lambert2024tulu} (\emph{pool \#1}), and three general-purpose language models (\emph{pool \#2}: \textsc{Qwen/Qwen2.5-7B-Instruct}, \textsc{meta-llama/Llama-3.1-8B-Instruct}, \textsc{deepseek-ai/DeepSeek-R1-Distill-Qwen-7B}).

We employ all four collaboration strategies in Section \ref{subsec:collaboration} (trained router, multiagent debate, logit fusion, and dare-ties) for model pool \#1, as they share the same tokenization and architecture, and model pool \#2 is only compatible with the first two model collaboration approaches. In trained router, we instantiate the router $f$ as \textsc{Qwen2.5-7B-Instruct} and train it with 1e-5 learning rate for 5 epochs; In multi-agent debate, we run the debate and run process three times for collaboration; In dare-ties merging, we employ the implementation of MergeKit \citep{goddard-etal-2024-arcees} and assign equal weight to models. We by default employ 512 max new tokens, temperature $\tau = 0.7$, and top-p $p=0.9$ sampling for text generation. We use the implementation of MoCo \citep{feng2026moco} for model collaboration algorithms.

We by default employ both supervised and multi-student knowledge distillation in Section \ref{subsec:distillation}, while we extend to on-policy distillation in Section \ref{sec:analysis}. We employ $\alpha:\beta:\gamma = 1:1:1$ for multi-student KD.

We by default execute the single-multi evolution loop for $k=3$ iterations and retain the best model/collaboration system based on the dev set.

\paragraph{Data and Evaluation} We employ 15 datasets spanning six domains for evaluation:
\begin{itemize}[leftmargin=*]
    \item QA: AGIEval \citep{zhong2024agieval}, ARC-challenge \citep{clark2018think}, MMLU-redux \citep{gema2025we}
    \item reasoning: BigBench-hard \citep{suzgun2023challenging}, GSM8k \citep{cobbe2021training}, MATH \citep{hendrycks2020measuring}, TheoremQA \citep{chen2023theoremqa}, GPQA-diamond \citep{rein2024gpqa}, TableMWP \citep{ludynamic})
    \item knowledge: WikiDYK \citep{zhang2025bidirectional}, PopQA \citep{mallen2023not}, BLEND \citep{myung2024blend}
    \item misc: safety (CocoNot \citep{brahman2024art}), science (Sciriff \citep{wadden2025sciriff}), and instruction following (human interest \citep{feng2025model})
\end{itemize}
We by default sample 1k data points for both dev and test sets for large datasets. We report both the $\mathrm{avg} \pm \mathrm{std}$ for the single models and the performance of the multi-model collaboration system.

\section{Results}
\label{sec:results}

We present the performance of the single-multi evolution loop with various model pools, collaboration strategies, and distillation settings in Table \ref{tab:big}.

\paragraph{The single-multi evolution loop is effective.} Averaging across datasets, collaboration strategies, and distillation settings, the single-multi evolution loops improve both individual models and the collaboration system they form by 8.0\% and 14.9\% on average. This indicates that models collectively evolve by interacting with an environment of other models and learning from the collaborative outcome.

\paragraph{Collaboration strategies such as routing and merging are more compatible.} In setting 1, API-level routing (53.39), weight-level model merging (52.72), text-level multiagent debate (50.29), and API-level logit fusion (42.86) achieved decreasing performance. We posit that the best aggregate of responses or model weights serves as a stronger teacher. This also indicates that the single-multi evolution loops's success depends on model collaboration strategies: we experiment with more collaboration strategies in Section \ref{sec:analysis}.

\paragraph{The single-multi evolution loop benefits from better distillation and model diversity.} Multi-student knowledge distillation on average outperforms simple supervised distillation by 4.67\% on average, indicating that the single-multi evolution loop benefits from stronger distillation methods and we further extend to on-policy distillation in Section \ref{sec:analysis}. In addition, the performance gains of API- and text-level approaches in model pool 1 (specialized LMs, 19.62\%) are higher than model pool 2 (general-purpose LMs, 10.03\%): while both decent, this indicates the single-multi evolution loop benefits from the complementary strength of  and specialized language models.

\paragraph{More effective on reasoning and knowledge.} On the three domains with more than one dataset, the single-multi evolution loops achieve 16.84\% (reasoning), 12.67\% (knowledge), and 7.52\% (QA) gains. This indicates that by empowering models to interact and collaborate with an environment of other models and learn from the collaborative outcome, models benefit from the complementary reasoning patterns and knowledge capacity of other models.

\begin{figure}[t]
    \centering
    \includegraphics[width=0.9\linewidth]{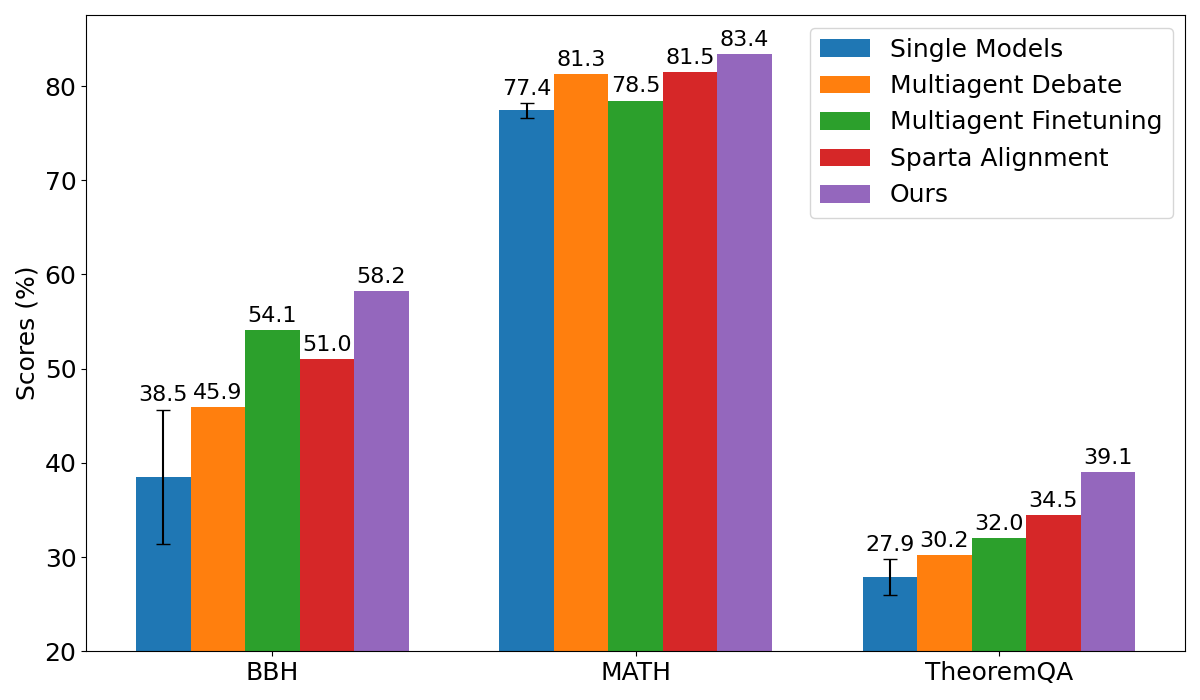}
    \caption{Comparing the single-multi evolution loop with existing evolution strategies. Our strategy consistently outperforms the three methods by 7.7\% on average.}
    \label{fig:evolution}
\end{figure}

\vspace*{-5pt}
\section{Analysis}
\label{sec:analysis}

\paragraph{Comparison with Other Evolution Methods} We compare the single-multi evolution loop with three existing methods of iterative evolution: multi-agent debate \citep{du2023improving}, where model responses are refined based on other models' responses; multi-agent fine-tuning \citep{subramaniammultiagent}, where models are trained collectively via critique and debate; sparta alignment \citep{jiang2025sparta}, where models evaluate and align each other under game theory settings. Figure \ref{fig:evolution} shows the performance under different evolution settings with model pool \#1 and three datasets: the single-multi evolution loop outperforms the three existing methods by 7.7\% on average, indicating that the strategy of introducing deeper collaboration and employing a collaborative system as the teacher for distillation yields better collective evolution of diverse language models.

\paragraph{On-Policy Distillation} Section \ref{sec:results} show that multi-student distillation generally outperforms simple supervised distillation, indicating that the single-multi evolution loop benefits from employing better distillation methods. We further extend this to on-policy distillation \citep{agarwal2024policy}, specifically GOLD \citep{patiño2025_unlocking_on_policy_distillation_for_any_model_family}, and employ dare-ties model merging the collaboration strategy $\mathcal{C}$ and model pool \#1.\footnotemark[3]\footnotetext[3]{This ensures that the teacher is a merged single model, compatible with the on-policy setting requiring logit information.} Results in Table \ref{tab:on-policy} show that on-policy distillation is even better than multi-student and supervised knowledge distillation, outperforming them by up to 5.2\%. We envision further exploration on the deep literature of knowledge distillation research, how that might benefit model collaboration and model evolution, and how to flexibly enable collaborative systems of multiple models as the teacher in distillation methods.

\begin{table}[t]
\centering
\scriptsize
\setlength{\tabcolsep}{1.2pt}
\renewcommand{\arraystretch}{1.3}
\caption{Employing on-policy distillation in the single-multi evolution loop yields further gains compared to multi-student and simple supervised distillation, indicating that the broad literature of knowledge distillation methods could be valuable for model collaboration and evolution.}
\vspace*{-5pt}
\label{tab:on-policy}
\resizebox{1\linewidth}{!}{ 
\begin{tabular}{lcccccccc}\toprule[1.5pt]
&\multicolumn{2}{c}{MMLU-redux} &\multicolumn{2}{c}{MATH} &\multicolumn{2}{c}{TheoremQA} &\multicolumn{2}{c}{BLEND} \\\cmidrule{2-9}
&Single &Multi &Single &Multi &Single &Multi &Single &Multi \\\midrule
Initial &65.57 &69.90 &77.41 &71.40 &27.92 &25.00 &72.23 &73.00 \\
Supervised KD &69.63 &73.90 &81.90 &77.30 &33.00 &27.75 &78.80 &74.85 \\
Multi-Student KD &71.67 &71.30 &82.04 &79.29 &34.33 &33.50 &78.40 &78.80 \\
On-Policy KD &\textbf{73.17} &\textbf{74.20} &\textbf{83.30} &\textbf{79.92} &\textbf{34.67} &\textbf{35.25} &\textbf{79.62} &\textbf{80.45} \\
\bottomrule[1.5pt]
\end{tabular}
}
\end{table}

\begin{figure}[t]
    \centering
    \vspace*{10pt}
    \includegraphics[width=1\linewidth]{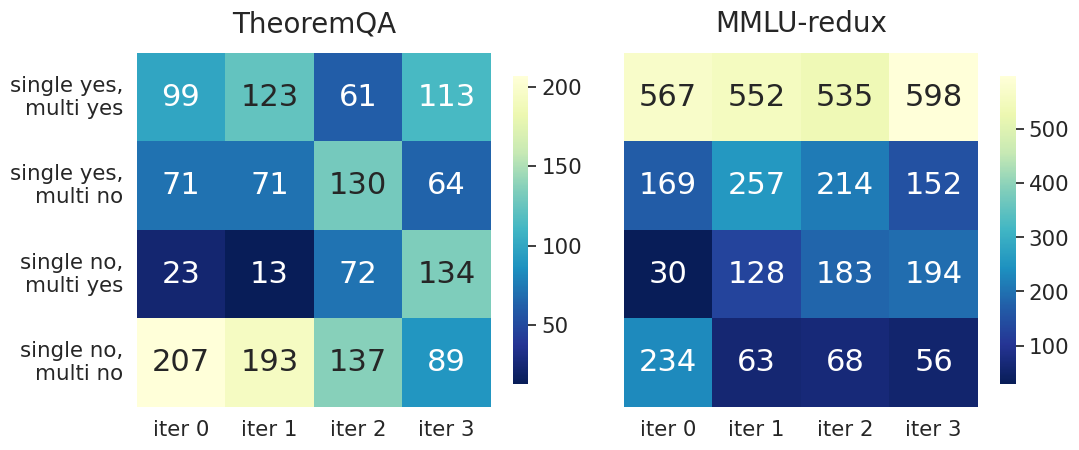}
    \vspace*{-5pt}
    \caption{How does the skills dynamic change through evolution iterations, \emph{i.e.} how many problems could be solved by at least one model individually or the multi-LLM collaboration system. Results indicate better and more synergistic models and systems.}
    \label{fig:skills}
\end{figure}

\begin{table}[t]
\centering
\scriptsize
\setlength{\tabcolsep}{1.5pt}
\renewcommand{\arraystretch}{1.3}
\caption{Results with three additional model collaboration strategies show that the single-multi evolution loop is a broadly effective idea compatible with diverse model collaboration algorithms.}
\vspace*{-5pt}
\label{tab:more_collaboration}
\resizebox{0.9\linewidth}{!}{ 
\begin{tabular}{lcccc}\toprule[1.5pt]
& &TableMWP &SciRiff \\\midrule[0.75pt]
\multicolumn{2}{c}{single, t=0} &58.93 (10.56) &51.80 (8.84) \\\midrule[0.75pt]
\multirow{3}{*}{LLM Blender} &multi, t=0 &78.40 &56.83 \\
&single, t=n &61.87 (10.00) &54.44 (6.12) \\
&multi, t=n &\textbf{80.00} &\textbf{58.99} \\
\multirow{3}{*}{Multiagent Finetuning} &multi, t=0 &76.00 &51.80 \\
&single, t=n &60.80 (9.23) &53.04 (7.76) \\
&multi, t=n &\underline{79.20} &56.12 \\
\multirow{3}{*}{Graph Routing} &multi, t=0 &64.80 &52.52 \\
&single, t=n &61.60 (12.87) &53.80 (8.11) \\
&multi, t=n &71.20 &\underline{58.27} \\
\bottomrule[1.5pt]
\end{tabular}
}
\end{table}

\begin{table}[t]
\centering
\scriptsize
\setlength{\tabcolsep}{1.5pt}
\renewcommand{\arraystretch}{1.3}
\caption{The single-multi evolution loop brings consistent improvements to the three official variants of \textsc{Qwen-2.5-7B}.}
\vspace*{-5pt}
\label{tab:more_models}
\resizebox{1\linewidth}{!}{ 
\begin{tabular}{lcccc}\toprule[1.5pt]
&MMLU-redux &MATH &TheoremQA &BLEND \\\midrule[0.75pt]
single, t=0 &54.70 (4.99) &75.77 (5.76) &29.08 (3.64) &58.33 (7.14) \\
multi, t=0 &48.70 &77.20 &31.00 &67.00 \\
single, t=n &\textbf{56.90} (5.28) &76.85 (4.83) &32.00 (2.78) &61.63 (6.49) \\
multi, t=n &52.50 &\textbf{80.86} &\textbf{33.50} &\textbf{69.30} \\
\bottomrule[1.5pt]
\end{tabular}
}
\end{table}

\paragraph{Skills Change of Single/Multi} Given an evaluation problem, whether the single models or the multi-LLM system can correctly solve the problem varies. We use ``single yes/no'' to denote whether any of the single models can individually solve the problem, and use ``multi yes/no'' to denote whether the model collaboration can do so. This creates a $2 \times 2$ confusion matrix highlighting the skill dynamics between individual models and the model collaboration system. We visualize the change of the skill dynamics across single-multi evolution iterations in Figure \ref{fig:skills}. 1) There are consistently fewer ``single no, multi no'' cases, indicating that the single-multi evolution loop helps models solve problems that were impossible to them initially; 2) There are consistently more ``single no, multi yes'' cases, indicating that the ceiling of collaboration is also raised due to more synergistic models through the collective evolution process.

\begin{table}[t]
\centering
\scriptsize
\setlength{\tabcolsep}{1.5pt}
\renewcommand{\arraystretch}{1.3}
\caption{The average and best performance of individual models, before and after evolution. This indicates that the single-multi evolution loop improves models both on average and on the best model's utility.}
\vspace*{-5pt}
\label{tab:best}
\resizebox{0.9\linewidth}{!}{ 
\begin{tabular}{lccc}\toprule[1.5pt]
&MATH &GPQA-dia &TheoremQA \\\midrule[0.75pt]
single, t=0, avg &72.97 (10.21) &25.93 (3.55) &27.08 (5.20) \\
single, t=0, best &81.69 &30.30 &32.75 \\
single, t=n, avg &75.84 (10.09) &29.63 (1.58) &29.17 (4.13) \\
single, t=n, best &84.26 &32.32 &35.00 \\
\bottomrule[1.5pt]
\end{tabular}
}
\end{table}


\begin{figure}[t]
    \centering
    \vspace*{10pt}
    \includegraphics[width=0.9\linewidth]{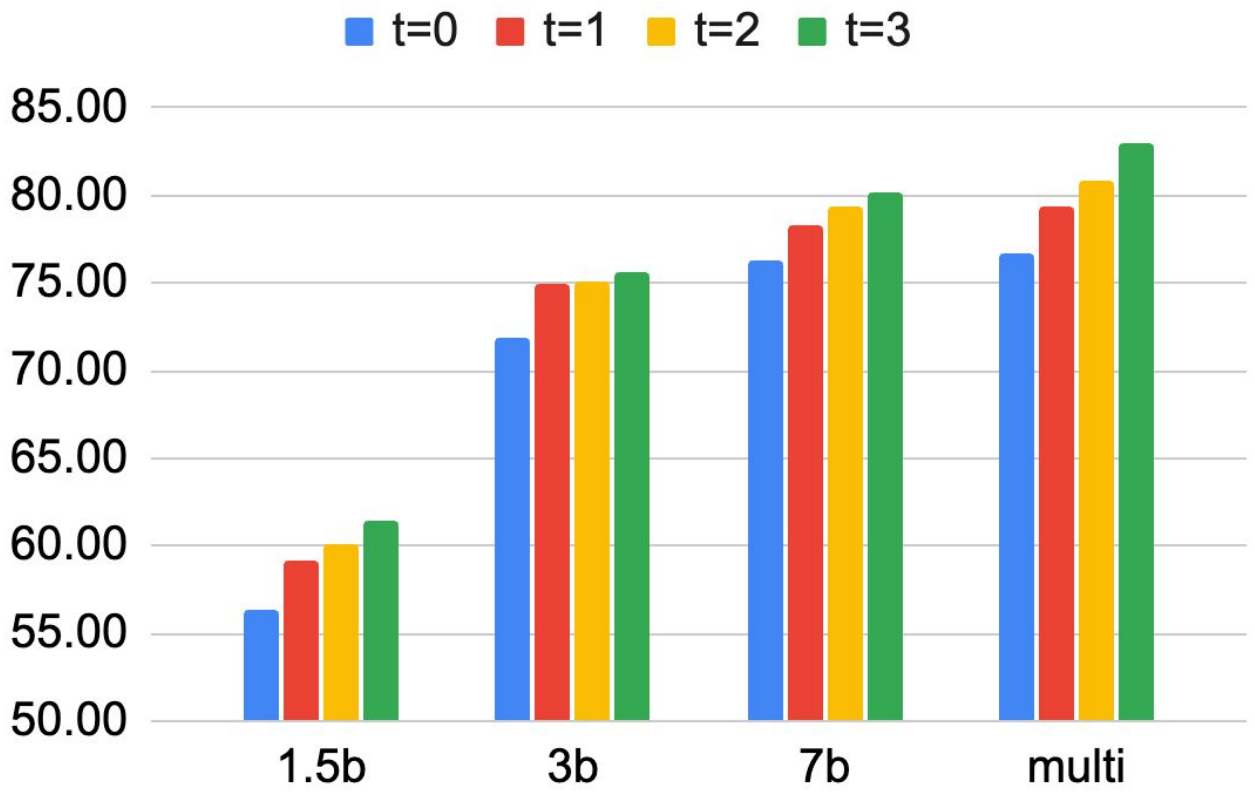}
    \caption{The performance of single individual models and multi-model systems across iterations, with a pool of three different sizes of \textsc{Qwen-2.5}. We employ multi-student KD and the BLEND dataset for evaluation.}
    \label{fig:size}
\end{figure}

\paragraph{More Collaboration Strategies $\mathcal{C}$} In Section \ref{sec:results}, we employ four model collaboration strategies to instantiate the single-multi evolution loop: trained router, multi-agent debate, logit fusion, and dare-ties model merging. We additionally test out the single-multi evolution loop with three more advanced model collaboration algorithms: LLM blender \citep{jiang2023llm}, where the responses of multiple LLMs and reranked and merged by training separate LMs; multi-agent fine-tuning \citep{subramaniammultiagent}, a training-based enhancement of multi-agent debate; and graph router \citep{fenggraphrouter}, employing graph neural networks as the routing strategy. Table \ref{tab:more_collaboration} shows that the single-multi evolution loop continues to bring improvements to both individual LMs and model collaboration systems with these three collaboration strategies, indicating its broad compatibility and effectiveness with diverse model collaboration protocols.

\begin{table*}[t]
    \centering
    \scriptsize
    \vspace*{5pt}
    \caption{Qualitative examples of model-generated responses to complex problems in the TheoremQA dataset.}
    \vspace*{5pt}
    \label{tab:qualitative}
    \resizebox{1\linewidth}{!}{ 
    \begin{tabularx}{1\textwidth}{l|X} 
        \toprule[1.5pt]
        \textbf{Model Version} & \textbf{Generated Text} \\
        \midrule[0.75pt]
        Question & Consider $x(t)$ to be given as, $ x(t)=\cos (1000 \pi t) $ . Let the sampling frequency be $2000 \mathrm{~Hz}$. Does aliasing occur? \\
        \addlinespace 
        Single, t=0 & Yes, aliasing will occur since the frequency of the signal is 1000 Hz which is equal to the Nyquist frequency, and the sampling frequency is also 2000 Hz which is twice the Nyquist frequency. Therefore, the signal cannot be accurately reconstructed from its samples due to aliasing. \xmark \\ \addlinespace
        Single, t=n & Step 1: Determine the Frequency of the Signal The given signal is: $x(t) = \cos(1000 \pi t)$. The general form of a cosine function is: $\cos(2 \pi f t)$. By comparing this with the given signal, we can see that: $2 \pi f = 1000 \pi$, which results in $f = 500 \text{ Hz}$. So, the frequency of the signal $x(t)$ is $500 \text{ Hz}$.
        
        Step 2: Determine the Nyquist Frequency The Nyquist frequency is half of the sampling frequency. Given the sampling frequency is $2000 \text{ Hz}$, the Nyquist frequency is: $f_{\text{Nyquist}} = \frac{2000 \text{ Hz}}{2} = 1000 \text{ Hz}$. 
        
        Step 3: Compare the Signal Frequency with the Nyquist Frequency For a signal to avoid aliasing, its frequency must be less than or equal to the Nyquist frequency. In this case: The signal frequency $f = 500 \text{ Hz}$ and the Nyquist frequency $f_{\text{Nyquist}} = 1000 \text{ Hz}$. Since $500 \text{ Hz} \leq 1000 \text{ Hz}$, the signal's frequency is within the non-aliasing range.
        
        Conclusion: No aliasing occurs because the signal frequency $500 \text{ Hz}$ is below the Nyquist frequency $1000 \text{ Hz}$. \cmark\\
        \midrule[0.75pt]
        Question & A cascade of $n$ identical independent binary symmetric channels each with raw error probability $p$, and $0 < p < 1$. What is the capacity of the cascade when $n$ goes to infinity? \\ \addlinespace
        Single, t=0 & The capacity of a cascade of $n$ identical independent binary symmetric channels each with raw error probability $p$ can be found using the following formula: $C = \lim_{n \to \infty} (1-p)^n \log_2(1-p) + p^n \log_2(p)$ where $C$ is the capacity of the cascade. To see why this formula holds, consider that as $n$ goes to infinity, the probability of no errors occurring in the cascade goes to $(1-p)^n$, and the probability of exactly one error occurring goes to $np^n$. All other probabilities go to zero. The capacity of the cascade is then given by the sum of the mutual information between the input and output of the cascade for each possible number of errors, weighted by their respective probabilities. This can be expressed as: $C = \sum_{k=0}^{\infty} P_k I(X;Y|k)$ where $P_k$ is the probability of exactly $k$ errors occurring, and $I(X;Y|k)$ is the mutual information between the input and output of the cascade conditioned on $k$ errors. Plugging in the expressions for $P_k$ and $I(X;Y|k)$ and simplifying, we obtain the formula above. \xmark \\ \addlinespace
        Single, t=n & To determine the capacity of a cascade of $n$ identical independent binary symmetric channels (BSCs) each with a raw error probability $p$, we need to understand how the error probability accumulates over the cascade as $n$ becomes large. Step-by-Step Reasoning:
        
        1. **Error Probability in a Single Channel:** - A BSC has a raw error probability $p$. - The probability of a correct transmission through a single channel is $1 - p$.
        
        2. **Probability of No Error in the Cascade:** - For $n$ independent channels, the probability that all $n$ channels transmit correctly is $(1 - p)^n$. - Therefore, the probability of at least one error occurring in the cascade is: $P(\text{at least one error}) = 1 - (1 - p)^n$
        
        3. **Capacity of the Cascade:** - The capacity $C$ of a channel is defined as the maximum rate at which information can be transmitted reliably. - For a BSC, the capacity is given by $C = 1 - H(p)$, where $H(p)$ is the binary entropy function: $H(p) = -p \log_2 p - (1 - p) \log_2 (1 - p)$ - In the context of the cascade, the capacity is influenced by the error probability of the cascade, not the individual channels.
        
        4. **Asymptotic Behavior as $n \to \infty$:** - As $n$ approaches infinity, the term $(1 - p)^n$ approaches 0 because $0 < p < 1$. - Therefore, the probability of no error in the cascade, $(1 - p)^n$, goes to 0. - Consequently, the probability of at least one error in the cascade, $1 - (1 - p)^n$, goes to 1.
        
        5. **Conclusion on Capacity:** - Since the probability of an error in the cascade approaches 1 as $n$ goes to infinity, the information cannot be transmitted reliably.   - Thus, the capacity of the cascade as $(n \to \infty)$ is 0. Final Answer: $0$ \cmark \\ \bottomrule[1.5pt]
    \end{tabularx}
    }
\end{table*}

\paragraph{Additional Model Pool} In Section \ref{sec:results}, we employ two pools of large language models by default: pool \#1 with three specialized LLMs and pool \#2 with three general-purpose LLMs. Here we additionally employ the three official variants of \textsc{Qwen-2.5-7B} (math, code, instruct) as an additional model pool and present results in Table \ref{tab:more_models}. The single-multi evolution loop again brings performance boosts to both individual models and the multi-LLM collaboration system, indicating its compatibility with diverse model settings.

We additionally test out a pool of models with different sizes: specifically \textsc{Qwen-2.5-1.5b}, \textsc{3b}, and \textsc{7b}. We investigate whether the single-multi evolution loop could bring improvements to models with various sizes through iterative collaboration and distillation.  Figure \ref{fig:size} demonstrates that the single-multi evolution loop is compatible with model collaboration of varying sizes, bringing performance gains of 6.44\% on average across participating models.

\paragraph{Average vs. Best} We by default report the $\mathrm{avg} \pm \mathrm{std}$ for single models, while the performance of the best single model in the pool is also a valuable indicator. We report the best model performance with multi-student knowledge distillation, the multi-agent debate collaboration strategy, and model pool \#2 in Table \ref{tab:best}. It shows that the single-multi evolution loop brings improvements to models not only \emph{on average}, but also raises the ceiling of the strongest model in the pool, boosting model utility in both collaborative and independent contexts.

\paragraph{Qualitative Examples} We present model-generated responses to complex questions in the TheoremQA dataset in Table \ref{tab:qualitative}. Despite the complexity of the scientific problem, evolved LLMs after the single-multi evolution loop generate significantly better and more informative responses compared to the pre-evolution models, leading to correct answers.

\section{Related Work}
\vspace*{10pt}

\paragraph{Model Collaboration} Aside from competing to develop the best language model, an increasing line of work is focusing on \emph{model collaboration} \citep{feng2025one}: multiple LLMs trained on different data by different people collaborate, compose, and complement each other in compositional AI systems. Model collaboration encompasses broad strokes of research and methods: API-level collaboration focuses on the routing \citep{ongroutellm, frick2025prompt, fenggraphrouter, zheng2025citer, llmrouter2025} and cascading \citep{chenfrugalgpt, guptalanguage, yuelarge} among multiple language models; text-level collaboration employs debate \citep{du2023improving, subramaniammultiagent}, feedback \citep{feng2024don}, response aggregation \citep{jiang2023llm, zhao2025majority}, mutual evaluation \citep{jiang2025sparta}, and more techniques based on the exchange of generated texts across models \citep{guo2024large, zhao2025language, dang2025multi, liang2024encouraging, zhao2024slide}; API-level collaboration features arithmetic and fusion on the token probabilities of language models \citep{pei2023preadd, li2023contrastive, mavromatispack, chuangdola, mitchellemulator, liutuning, huangdivide}; weight-level collaboration features model merging \citep{yadav2023ties, yu2024language}, model arithmetic \citep{huanglorahub, feng2025model, zheng2024weak}, and Mixture-of-Experts \citep{yadavsurvey, shi2025flexolmo}.

In this work, we seek to improve the efficiency of multi-LLM collaboration systems by \emph{distilling} their collaboratively generated texts back into a single language model, reducing the inference-time cost to only using a single LLM while imitating the collaboration patterns and skills of multiple models. We further show that the post-distillation improved language models could collaborate again to form stronger model collaboration systems, and we term this iterative collaboration-distillation evolution process ``the single-multi evolution loop''.

\paragraph{Knowledge Distillation} Knowledge distillation \citep{hinton2015distillingknowledgeneuralnetwork, sanh2020distilbertdistilledversionbert, yang2024qwen2technicalreport} employs a pair of student and teacher machine learning models and aims to transfer the skills of the larger teacher model to the smaller and often weaker student model. Early distillation methods directly employ the teacher-generated outputs as ground truths for students to learn from via supervised fine-tuning \citep{kim-rush-2016-sequence, alpaca, do2025effectiveness}. To obtain finer-grained signals, the logit distributions of the teacher model could serve as a better signal for soft-label distillation \citep{lin2020autoregressive, gu2024minillmknowledgedistillationlarge}. Recent work proposed on-policy distillation and variants \citep{agarwal2024policy, ko2024distillmstreamlineddistillationlarge, xuspeculative, patiño2025_unlocking_on_policy_distillation_for_any_model_family}, seeking to also leverage the generated tokens of the student model to bridge the skill gap between the student and teacher models and mitigate memorization.

In this work, we experiment with a new knowledge distillation setting, where the teacher is no longer a single language model, but a model collaboration system of multiple language models. We aim to distill the collaborative patterns and outcomes of expensive model collaboration systems back into a single language model for inference-time efficiency, and these distilled and improved LLMs could collaborate again to form stronger model collaboration systems. This relates to distillation specifically in multi-agent settings \citep{chen2024magdi} (text-level collaboration in our definition), while we focus on the broader spectrum of model collaboration algorithms. We envision future work in the distillation domain specifically tailored to employing modular and compositional AI systems as teachers or students in knowledge distillation.


\section{Conclusion}
We propose the \emph{single-multi evolution loop}, where LLMs collaborate and interact with an environment of other models for collective evolution. In the multi-step, LLMs collaborate via model collaboration algorithms to form stronger systems. In the single-step, each LLM then distills from the collaborative outputs of the multi-LLM system to learn from the collaborative patterns. These stronger post-distillation LLMs then collaborate again in the multi-step to form even stronger model collaboration systems. By alternating between collaboration and distillation in the single-multi evolution loop, multiple LLMs form a collective evolution ecosystem where models evolve by interacting with an environment of other models. We conduct extensive experiments with 15 datasets, seven collaboration strategies, three distillation methods, and three model pools: the single-multi evolution loop brings 8.0\% and 14.9\% average improvements to both individual LMs and the multi-LM systems. Further analysis reveals that it outperforms existing multi-LLM evolution strategies and enhances the skills of models and systems to solve problems initially beyond their capability.

\section*{Impact Statement}
The single-multi evolution loop is a strong strategy to collectively improve multiple language models. When deployed to the wild and when the multiple models are contributed by different people, we envision potential safety risks if one of the models is malicious/compromised. We suggest that users should take extra caution when selecting models for the single-multi evolution loop and envisioning future work on safeguarding collaborative and evolutionary AI systems from malicious contributions.

\bibliography{example_paper, distill}

@misc{ko2024distillmstreamlineddistillationlarge,
      title={DistiLLM: Towards Streamlined Distillation for Large Language Models}, 
      author={Jongwoo Ko and Sungnyun Kim and Tianyi Chen and Se-Young Yun},
      year={2024},
      eprint={2402.03898},
      archivePrefix={arXiv},
      primaryClass={cs.CL},
}

@article{kim2016sequence,
  title={Sequence-level knowledge distillation},
  author={Kim, Yoon and Rush, Alexander M},
  journal={arXiv preprint arXiv:1606.07947},
  year={2016}
}

@misc{hinton2015distillingknowledgeneuralnetwork,
      title={Distilling the Knowledge in a Neural Network}, 
      author={Geoffrey Hinton and Oriol Vinyals and Jeff Dean},
      year={2015},
      eprint={1503.02531},
      archivePrefix={arXiv},
      primaryClass={stat.ML},
}

@misc{sanh2020distilbertdistilledversionbert,
      title={DistilBERT, a distilled version of BERT: smaller, faster, cheaper and lighter}, 
      author={Victor Sanh and Lysandre Debut and Julien Chaumond and Thomas Wolf},
      year={2020},
      eprint={1910.01108},
      archivePrefix={arXiv},
      primaryClass={cs.CL},
}

@inproceedings{kim-rush-2016-sequence,
    title = "Sequence-Level Knowledge Distillation",
    author = "Kim, Yoon  and
      Rush, Alexander M.",
    editor = "Su, Jian  and
      Duh, Kevin  and
      Carreras, Xavier",
    booktitle = "Proceedings of the 2016 Conference on Empirical Methods in Natural Language Processing",
    month = nov,
    year = "2016",
    address = "Austin, Texas",
    publisher = "Association for Computational Linguistics",
    doi = "10.18653/v1/D16-1139",
    pages = "1317--1327",
}

@misc{alpaca,
  author = {Rohan Taori and Ishaan Gulrajani and Tianyi Zhang and Yann Dubois and Xuechen Li and Carlos Guestrin and Percy Liang and Tatsunori B. Hashimoto },
  title = {Stanford Alpaca: An Instruction-following LLaMA model},
  year = {2023},
  publisher = {GitHub},
  journal = {GitHub repository},
  howpublished = {\url{https://github.com/tatsu-lab/stanford_alpaca}},
}

@misc{gu2024minillmknowledgedistillationlarge,
      title={MiniLLM: Knowledge Distillation of Large Language Models}, 
      author={Yuxian Gu and Li Dong and Furu Wei and Minlie Huang},
      year={2024},
      eprint={2306.08543},
      archivePrefix={arXiv},
      primaryClass={cs.CL},
}

@article{liu2024skywork,
  title={Skywork-reward: Bag of tricks for reward modeling in llms},
  author={Liu, Chris Yuhao and Zeng, Liang and Liu, Jiacai and Yan, Rui and He, Jujie and Wang, Chaojie and Yan, Shuicheng and Liu, Yang and Zhou, Yahui},
  journal={arXiv preprint arXiv:2410.18451},
  year={2024}
}

@misc{yang2024qwen2technicalreport,
      title={Qwen2 Technical Report}, 
      author={An Yang and Baosong Yang and Binyuan Hui and Bo Zheng and Bowen Yu and Chang Zhou and Chengpeng Li and Chengyuan Li and Dayiheng Liu and Fei Huang and Guanting Dong and Haoran Wei and Huan Lin and Jialong Tang and Jialin Wang and Jian Yang and Jianhong Tu and Jianwei Zhang and Jianxin Ma and Jianxin Yang and Jin Xu and Jingren Zhou and Jinze Bai and Jinzheng He and Junyang Lin and Kai Dang and Keming Lu and Keqin Chen and Kexin Yang and Mei Li and Mingfeng Xue and Na Ni and Pei Zhang and Peng Wang and Ru Peng and Rui Men and Ruize Gao and Runji Lin and Shijie Wang and Shuai Bai and Sinan Tan and Tianhang Zhu and Tianhao Li and Tianyu Liu and Wenbin Ge and Xiaodong Deng and Xiaohuan Zhou and Xingzhang Ren and Xinyu Zhang and Xipin Wei and Xuancheng Ren and Xuejing Liu and Yang Fan and Yang Yao and Yichang Zhang and Yu Wan and Yunfei Chu and Yuqiong Liu and Zeyu Cui and Zhenru Zhang and Zhifang Guo and Zhihao Fan},
      year={2024},
      eprint={2407.10671},
      archivePrefix={arXiv},
      primaryClass={cs.CL}, 
}

@inproceedings{xuspeculative,
  title={Speculative Knowledge Distillation: Bridging the Teacher-Student Gap Through Interleaved Sampling},
  author={Xu, Wenda and Han, Rujun and Wang, Zifeng and Le, Long and Madeka, Dhruv and Li, Lei and Wang, William Yang and Agarwal, Rishabh and Lee, Chen-Yu and Pfister, Tomas},
  booktitle={The Thirteenth International Conference on Learning Representations},
  year={2025}
}

@misc{do2025effectiveness,
      title={Effectiveness of Chain-of-Thought in Distilling Reasoning Capability from Large Language Models}, 
      author={Cong-Thanh Do and Rama Doddipatla and Kate Knill},
      year={2025},
      eprint={2511.05184},
      archivePrefix={arXiv},
      primaryClass={cs.CL}
}

@inproceedings{chen2024magdi,
  title={MAGDi: Structured Distillation of Multi-Agent Interaction Graphs Improves Reasoning in Smaller Language Models},
  author={Chen, Justin and Saha, Swarnadeep and Stengel-Eskin, Elias and Bansal, Mohit},
  booktitle={International Conference on Machine Learning},
  pages={7220--7235},
  year={2024},
  organization={PMLR}
}

@article{feng2026moco,
  title={MoCo: A One-Stop Shop for Model Collaboration Research},
  author={Feng, Shangbin and Bai, Yuyang and Yang, Ziyuan and Wang, Yike and Tan, Zhaoxuan and Yan, Jiajie and Lei, Zhenyu and Ding, Wenxuan and Shi, Weijia and Wang, Haojin and others},
  journal={arXiv preprint arXiv:2601.21257},
  year={2026}
}

@article{feng2025one,
  title={When one llm drools, multi-llm collaboration rules},
  author={Feng, Shangbin and Ding, Wenxuan and Liu, Alisa and Wang, Zifeng and Shi, Weijia and Wang, Yike and Shen, Zejiang and Han, Xiaochuang and Lang, Hunter and Lee, Chen-Yu and others},
  journal={arXiv preprint arXiv:2502.04506},
  year={2025}
}

@inproceedings{ongroutellm,
  title={RouteLLM: Learning to Route LLMs from Preference Data},
  author={Ong, Isaac and Almahairi, Amjad and Wu, Vincent and Chiang, Wei-Lin and Wu, Tianhao and Gonzalez, Joseph E and Kadous, M Waleed and Stoica, Ion},
  booktitle={The Thirteenth International Conference on Learning Representations},
  year={2025}
}

@article{frick2025prompt,
  title={Prompt-to-leaderboard},
  author={Frick, Evan and Chen, Connor and Tennyson, Joseph and Li, Tianle and Chiang, Wei-Lin and Angelopoulos, Anastasios N and Stoica, Ion},
  journal={arXiv preprint arXiv:2502.14855},
  year={2025}
}

@inproceedings{shen2024learning,
  title={Learning to Decode Collaboratively with Multiple Language Models},
  author={Shen, Zejiang and Lang, Hunter and Wang, Bailin and Kim, Yoon and Sontag, David},
  booktitle={Proceedings of the 62nd Annual Meeting of the Association for Computational Linguistics (Volume 1: Long Papers)},
  pages={12974--12990},
  year={2024}
}

@article{lambert2024tulu,
  title={Tulu 3: Pushing frontiers in open language model post-training},
  author={Lambert, Nathan and Morrison, Jacob and Pyatkin, Valentina and Huang, Shengyi and Ivison, Hamish and Brahman, Faeze and Miranda, Lester James V and Liu, Alisa and Dziri, Nouha and Lyu, Shane and others},
  journal={arXiv preprint arXiv:2411.15124},
  year={2024}
}

@inproceedings{du2023improving,
  title={Improving factuality and reasoning in language models through multiagent debate},
  author={Du, Yilun and Li, Shuang and Torralba, Antonio and Tenenbaum, Joshua B and Mordatch, Igor},
  booktitle={Forty-first International Conference on Machine Learning},
  year={2023}
}

@inproceedings{liutuning,
  title={Tuning Language Models by Proxy},
  author={Liu, Alisa and Han, Xiaochuang and Wang, Yizhong and Tsvetkov, Yulia and Choi, Yejin and Smith, Noah A},
  booktitle={First Conference on Language Modeling},
  year={2024}
}

@article{yadav2023ties,
  title={Ties-merging: Resolving interference when merging models},
  author={Yadav, Prateek and Tam, Derek and Choshen, Leshem and Raffel, Colin A and Bansal, Mohit},
  journal={Advances in Neural Information Processing Systems},
  volume={36},
  pages={7093--7115},
  year={2023}
}

@inproceedings{yu2024language,
  title={Language models are super mario: Absorbing abilities from homologous models as a free lunch},
  author={Yu, Le and Yu, Bowen and Yu, Haiyang and Huang, Fei and Li, Yongbin},
  booktitle={Forty-first International Conference on Machine Learning},
  year={2024}
}

@article{zhang2025bidirectional,
  title={Bidirectional lms are better knowledge memorizers? a benchmark for real-world knowledge injection},
  author={Zhang, Yuwei and Yu, Wenhao and Feng, Shangbin and Zhu, Yifan and Peng, Letian and Srinivasa, Jayanth and Liu, Gaowen and Shang, Jingbo},
  journal={arXiv preprint arXiv:2505.12306},
  year={2025}
}

@inproceedings{lin2022truthfulqa,
  title={TruthfulQA: Measuring How Models Mimic Human Falsehoods},
  author={Lin, Stephanie and Hilton, Jacob and Evans, Owain},
  booktitle={Proceedings of the 60th Annual Meeting of the Association for Computational Linguistics (Volume 1: Long Papers)},
  pages={3214--3252},
  year={2022}
}

@inproceedings{feng2024don,
  title={Don’t Hallucinate, Abstain: Identifying LLM Knowledge Gaps via Multi-LLM Collaboration},
  author={Feng, Shangbin and Shi, Weijia and Wang, Yike and Ding, Wenxuan and Balachandran, Vidhisha and Tsvetkov, Yulia},
  booktitle={Proceedings of the 62nd Annual Meeting of the Association for Computational Linguistics (Volume 1: Long Papers)},
  pages={14664--14690},
  year={2024}
}

@article{cobbe2021training,
  title={Training verifiers to solve math word problems},
  author={Cobbe, Karl and Kosaraju, Vineet and Bavarian, Mohammad and Chen, Mark and Jun, Heewoo and Kaiser, Lukasz and Plappert, Matthias and Tworek, Jerry and Hilton, Jacob and Nakano, Reiichiro and others},
  journal={arXiv preprint arXiv:2110.14168},
  year={2021}
}

@inproceedings{suzgun2023challenging,
  title={Challenging BIG-Bench Tasks and Whether Chain-of-Thought Can Solve Them},
  author={Suzgun, Mirac and Scales, Nathan and Sch{\"a}rli, Nathanael and Gehrmann, Sebastian and Tay, Yi and Chung, Hyung Won and Chowdhery, Aakanksha and Le, Quoc and Chi, Ed and Zhou, Denny and others},
  booktitle={Findings of the Association for Computational Linguistics: ACL 2023},
  pages={13003--13051},
  year={2023}
}

@article{brahman2024art,
  title={The art of saying no: Contextual noncompliance in language models},
  author={Brahman, Faeze and Kumar, Sachin and Balachandran, Vidhisha and Dasigi, Pradeep and Pyatkin, Valentina and Ravichander, Abhilasha and Wiegreffe, Sarah and Dziri, Nouha and Chandu, Khyathi and Hessel, Jack and others},
  journal={Advances in Neural Information Processing Systems},
  volume={37},
  pages={49706--49748},
  year={2024}
}

@inproceedings{zhong2024agieval,
  title={AGIEval: A Human-Centric Benchmark for Evaluating Foundation Models},
  author={Zhong, Wanjun and Cui, Ruixiang and Guo, Yiduo and Liang, Yaobo and Lu, Shuai and Wang, Yanlin and Saied, Amin and Chen, Weizhu and Duan, Nan},
  booktitle={Findings of the Association for Computational Linguistics: NAACL 2024},
  pages={2299--2314},
  year={2024}
}

@inproceedings{mallen2023not,
  title={When Not to Trust Language Models: Investigating Effectiveness of Parametric and Non-Parametric Memories},
  author={Mallen, Alex and Asai, Akari and Zhong, Victor and Das, Rajarshi and Khashabi, Daniel and Hajishirzi, Hannaneh},
  booktitle={Proceedings of the 61st Annual Meeting of the Association for Computational Linguistics (Volume 1: Long Papers)},
  pages={9802--9822},
  year={2023}
}

@article{jiang2025sparta,
  title={SPARTA ALIGNMENT: Collectively Aligning Multiple Language Models through Combat},
  author={Jiang, Yuru and Ding, Wenxuan and Feng, Shangbin and Durrett, Greg and Tsvetkov, Yulia},
  journal={arXiv preprint arXiv:2506.04721},
  year={2025}
}

@article{clark2018think,
  title={Think you have solved question answering? try arc, the ai2 reasoning challenge},
  author={Clark, Peter and Cowhey, Isaac and Etzioni, Oren and Khot, Tushar and Sabharwal, Ashish and Schoenick, Carissa and Tafjord, Oyvind},
  journal={arXiv preprint arXiv:1803.05457},
  year={2018}
}

@inproceedings{hendrycks2020measuring,
  title={Measuring massive multitask language understanding},
  author={Hendrycks, Dan and Burns, Collin and Basart, Steven and Zou, Andy and Mazeika, Mantas and Song, Dawn and Steinhardt, Jacob},
  booktitle={International Conference on Learning Representations},
  year={2021}
}

@inproceedings{feng2025model,
  title={Model swarms: Collaborative search to adapt llm experts via swarm intelligence},
  author={Feng, Shangbin and Wang, Zifeng and Wang, Yike and Ebrahimi, Sayna and Palangi, Hamid and Miculicich, Lesly and Kulshrestha, Achin and Rauschmayr, Nathalie and Choi, Yejin and Tsvetkov, Yulia and others},
  booktitle={Forty-second International Conference on Machine Learning},
  year={2025}
}

@inproceedings{yuelarge,
  title={Large Language Model Cascades with Mixture of Thought Representations for Cost-Efficient Reasoning},
  author={Yue, Murong and Zhao, Jie and Zhang, Min and Du, Liang and Yao, Ziyu},
  booktitle={The Twelfth International Conference on Learning Representations},
  year={2024}
}

@article{chenfrugalgpt,
  title={FrugalGPT: How to Use Large Language Models While Reducing Cost and Improving Performance},
  author={Chen, Lingjiao and Zaharia, Matei and Zou, James},
  journal={Transactions on Machine Learning Research},
  year={2023}
}

@article{yadavsurvey,
  title={A Survey on Model MoErging: Recycling and Routing Among Specialized Experts for Collaborative Learning},
  author={Yadav, Prateek and Raffel, Colin and Muqeeth, Mohammed and Caccia, Lucas and Liu, Haokun and Chen, Tianlong and Bansal, Mohit and Choshen, Leshem and Sordoni, Alessandro},
  journal={Transactions on Machine Learning Research},
  year={2024}
}

@article{shi2025flexolmo,
  title={FlexOlmo: Open Language Models for Flexible Data Use},
  author={Shi, Weijia and Bhagia, Akshita and Farhat, Kevin and Muennighoff, Niklas and Walsh, Pete and Morrison, Jacob and Schwenk, Dustin and Longpre, Shayne and Poznanski, Jake and Ettinger, Allyson and others},
  journal={arXiv preprint arXiv:2507.07024},
  year={2025}
}

@inproceedings{fenggraphrouter,
  title={GraphRouter: A Graph-based Router for LLM Selections},
  author={Feng, Tao and Shen, Yanzhen and You, Jiaxuan},
  booktitle={The Thirteenth International Conference on Learning Representations},
  year={2025}
}

@inproceedings{guptalanguage,
  title={Language Model Cascades: Token-Level Uncertainty And Beyond},
  author={Gupta, Neha and Narasimhan, Harikrishna and Jitkrittum, Wittawat and Rawat, Ankit Singh and Menon, Aditya Krishna and Kumar, Sanjiv},
  booktitle={The Twelfth International Conference on Learning Representations},
  year={2024}
}

@inproceedings{guo2024large,
  title={Large language model based multi-agents: a survey of progress and challenges},
  author={Guo, Taicheng and Chen, Xiuying and Wang, Yaqi and Chang, Ruidi and Pei, Shichao and Chawla, Nitesh V and Wiest, Olaf and Zhang, Xiangliang},
  booktitle={Proceedings of the Thirty-Third International Joint Conference on Artificial Intelligence},
  pages={8048--8057},
  year={2024}
}

@inproceedings{liang2024encouraging,
  title={Encouraging Divergent Thinking in Large Language Models through Multi-Agent Debate},
  author={Liang, Tian and He, Zhiwei and Jiao, Wenxiang and Wang, Xing and Wang, Yan and Wang, Rui and Yang, Yujiu and Shi, Shuming and Tu, Zhaopeng},
  booktitle={Proceedings of the 2024 Conference on Empirical Methods in Natural Language Processing},
  pages={17889--17904},
  year={2024}
}

@inproceedings{zhao2025language,
  title={Language Model Council: Democratically Benchmarking Foundation Models on Highly Subjective Tasks},
  author={Zhao, Justin and Plaza-Del-Arco, Flor Miriam and Curry, Amanda Cercas},
  booktitle={Proceedings of the 2025 Conference of the Nations of the Americas Chapter of the Association for Computational Linguistics: Human Language Technologies (Volume 1: Long Papers)},
  pages={12395--12450},
  year={2025}
}

@inproceedings{li2023contrastive,
  title={Contrastive Decoding: Open-ended Text Generation as Optimization},
  author={Li, Xiang Lisa and Holtzman, Ari and Fried, Daniel and Liang, Percy and Eisner, Jason and Hashimoto, Tatsunori B and Zettlemoyer, Luke and Lewis, Mike},
  booktitle={Proceedings of the 61st Annual Meeting of the Association for Computational Linguistics (Volume 1: Long Papers)},
  pages={12286--12312},
  year={2023}
}

@inproceedings{mitchellemulator,
  title={An Emulator for Fine-tuning Large Language Models using Small Language Models},
  author={Mitchell, Eric and Rafailov, Rafael and Sharma, Archit and Finn, Chelsea and Manning, Christopher D},
  booktitle={The Twelfth International Conference on Learning Representations},
  year={2024}
}

@inproceedings{zhao2024slide,
  title={SLIDE: A Framework Integrating Small and Large Language Models for Open-Domain Dialogues Evaluation},
  author={Zhao, Kun and Yang, Bohao and Tang, Chen and Lin, Chenghua and Zhan, Liang},
  booktitle={Findings of the Association for Computational Linguistics ACL 2024},
  pages={15421--15435},
  year={2024}
}

@inproceedings{mavromatispack,
  title={Pack of LLMs: Model Fusion at Test-Time via Perplexity Optimization},
  author={Mavromatis, Costas and Karypis, Petros and Karypis, George},
  booktitle={First Conference on Language Modeling},
  year={2024}
}

@inproceedings{pei2023preadd,
  title={PREADD: Prefix-Adaptive Decoding for Controlled Text Generation},
  author={Pei, Jonathan and Yang, Kevin and Klein, Dan},
  booktitle={Findings of the Association for Computational Linguistics: ACL 2023},
  pages={10018--10037},
  year={2023}
}

@inproceedings{chuangdola,
  title={DoLa: Decoding by Contrasting Layers Improves Factuality in Large Language Models},
  author={Chuang, Yung-Sung and Xie, Yujia and Luo, Hongyin and Kim, Yoon and Glass, James R and He, Pengcheng},
  booktitle={The Twelfth International Conference on Learning Representations},
  year={2024}
}

@article{zheng2025citer,
  title={Citer: Collaborative inference for efficient large language model decoding with token-level routing},
  author={Zheng, Wenhao and Chen, Yixiao and Zhang, Weitong and Kundu, Souvik and Li, Yun and Liu, Zhengzhong and Xing, Eric P and Wang, Hongyi and Yao, Huaxiu},
  journal={arXiv preprint arXiv:2502.01976},
  year={2025}
}

@inproceedings{huangdivide,
  title={Divide, Reweight, and Conquer: A Logit Arithmetic Approach for In-Context Learning},
  author={Huang, Chengsong and Huang, Langlin and Huang, Jiaxin},
  booktitle={Workshop on Reasoning and Planning for Large Language Models},
  year={2025}
}

@inproceedings{huanglorahub,
  title={LoraHub: Efficient Cross-Task Generalization via Dynamic LoRA Composition},
  author={Huang, Chengsong and Liu, Qian and Lin, Bill Yuchen and Pang, Tianyu and Du, Chao and Lin, Min},
  booktitle={First Conference on Language Modeling},
  year={2024}
}

@inproceedings{goddard-etal-2024-arcees,
    title = "Arcee{'}s {M}erge{K}it: A Toolkit for Merging Large Language Models",
    author = "Goddard, Charles  and
      Siriwardhana, Shamane  and
      Ehghaghi, Malikeh  and
      Meyers, Luke  and
      Karpukhin, Vladimir  and
      Benedict, Brian  and
      McQuade, Mark  and
      Solawetz, Jacob",
    editor = "Dernoncourt, Franck  and
      Preo{\c{t}}iuc-Pietro, Daniel  and
      Shimorina, Anastasia",
    booktitle = "Proceedings of the 2024 Conference on Empirical Methods in Natural Language Processing: Industry Track",
    month = nov,
    year = "2024",
}

@article{dang2025multi,
  title={Multi-Agent Collaboration via Evolving Orchestration},
  author={Dang, Yufan and Qian, Chen and Luo, Xueheng and Fan, Jingru and Xie, Zihao and Shi, Ruijie and Chen, Weize and Yang, Cheng and Che, Xiaoyin and Tian, Ye and others},
  journal={arXiv preprint arXiv:2505.19591},
  year={2025}
}

@inproceedings{lin2020autoregressive,
  title={Autoregressive Knowledge Distillation through Imitation Learning},
  author={Lin, Alexander and Wohlwend, Jeremy and Chen, Howard and Lei, Tao},
  booktitle={Proceedings of the 2020 Conference on Empirical Methods in Natural Language Processing (EMNLP)},
  pages={6121--6133},
  year={2020}
}

@inproceedings{agarwal2024policy,
  title={On-policy distillation of language models: Learning from self-generated mistakes},
  author={Agarwal, Rishabh and Vieillard, Nino and Zhou, Yongchao and Stanczyk, Piotr and Garea, Sabela Ramos and Geist, Matthieu and Bachem, Olivier},
  booktitle={The twelfth international conference on learning representations},
  year={2024}
}

@inproceedings{hulora,
  title={LoRA: Low-Rank Adaptation of Large Language Models},
  author={Hu, Edward J and Wallis, Phillip and Allen-Zhu, Zeyuan and Li, Yuanzhi and Wang, Shean and Wang, Lu and Chen, Weizhu and others},
  booktitle={International Conference on Learning Representations},
  year={2022}
}

@inproceedings{lei2025learning,
  title={Learning from diverse reasoning paths with routing and collaboration},
  author={Lei, Zhenyu and Tan, Zhen and Wang, Song and Zhu, Yaochen and Chen, Zihan and Dong, Yushun and Li, Jundong},
  booktitle={Proceedings of the 2025 Conference on Empirical Methods in Natural Language Processing},
  pages={2832--2845},
  year={2025}
}

@misc{patiño2025_unlocking_on_policy_distillation_for_any_model_family,
  title={Unlocking On-Policy Distillation for Any Model Family},
  author={Carlos Miguel Patiño and Kashif Rasul and Quentin Gallouédec and Ben Burtenshaw and Sergio Paniego and Vaibhav Srivastav and Thibaud Frere and Ed Beeching and Lewis Tunstall and Leandro von Werra and Thomas Wolf},
  year={2025},
}

@inproceedings{gema2025we,
  title={Are we done with mmlu?},
  author={Gema, Aryo Pradipta and Leang, Joshua Ong Jun and Hong, Giwon and Devoto, Alessio and Mancino, Alberto Carlo Maria and Saxena, Rohit and He, Xuanli and Zhao, Yu and Du, Xiaotang and Madani, Mohammad Reza Ghasemi and others},
  booktitle={Proceedings of the 2025 Conference of the Nations of the Americas Chapter of the Association for Computational Linguistics: Human Language Technologies (Volume 1: Long Papers)},
  pages={5069--5096},
  year={2025}
}

@inproceedings{chen2023theoremqa,
  title={TheoremQA: A Theorem-driven Question Answering Dataset},
  author={Chen, Wenhu and Yin, Ming and Ku, Max and Lu, Pan and Wan, Yixin and Ma, Xueguang and Xu, Jianyu and Wang, Xinyi and Xia, Tony},
  booktitle={Proceedings of the 2023 Conference on Empirical Methods in Natural Language Processing},
  pages={7889--7901},
  year={2023}
}

@inproceedings{rein2024gpqa,
  title={Gpqa: A graduate-level google-proof q\&a benchmark},
  author={Rein, David and Hou, Betty Li and Stickland, Asa Cooper and Petty, Jackson and Pang, Richard Yuanzhe and Dirani, Julien and Michael, Julian and Bowman, Samuel R},
  booktitle={First Conference on Language Modeling},
  year={2024}
}

@inproceedings{ludynamic,
  title={Dynamic Prompt Learning via Policy Gradient for Semi-structured Mathematical Reasoning},
  author={Lu, Pan and Qiu, Liang and Chang, Kai-Wei and Wu, Ying Nian and Zhu, Song-Chun and Rajpurohit, Tanmay and Clark, Peter and Kalyan, Ashwin},
  booktitle={The Eleventh International Conference on Learning Representations},
  year={2023}
}

@article{myung2024blend,
  title={Blend: A benchmark for llms on everyday knowledge in diverse cultures and languages},
  author={Myung, Junho and Lee, Nayeon and Zhou, Yi and Jin, Jiho and Putri, Rifki and Antypas, Dimosthenis and Borkakoty, Hsuvas and Kim, Eunsu and Perez-Almendros, Carla and Ayele, Abinew Ali and others},
  journal={Advances in Neural Information Processing Systems},
  volume={37},
  pages={78104--78146},
  year={2024}
}

@inproceedings{wadden2025sciriff,
  title={Sciriff: A resource to enhance language model instruction-following over scientific literature},
  author={Wadden, David and Shi, Kejian and Morrison, Jacob and Li, Alan and Naik, Aakanksha and Singh, Shruti and Barzilay, Nitzan and Lo, Kyle and Hope, Tom and Soldaini, Luca and others},
  booktitle={Proceedings of the 2025 Conference on Empirical Methods in Natural Language Processing},
  pages={6083--6120},
  year={2025}
}

@inproceedings{subramaniammultiagent,
  title={Multiagent Finetuning: Self Improvement with Diverse Reasoning Chains},
  author={Subramaniam, Vighnesh and Du, Yilun and Tenenbaum, Joshua B and Torralba, Antonio and Li, Shuang and Mordatch, Igor},
  booktitle={The Thirteenth International Conference on Learning Representations},
  year={2025}
}

@inproceedings{jiang2023llm,
  title={LLM-Blender: Ensembling Large Language Models with Pairwise Ranking and Generative Fusion},
  author={Jiang, Dongfu and Ren, Xiang and Lin, Bill Yuchen},
  booktitle={Proceedings of the 61st Annual Meeting of the Association for Computational Linguistics (Volume 1: Long Papers)},
  pages={14165--14178},
  year={2023}
}

@misc{llmrouter2025,
  title        = {LLMRouter: An Open-Source Library for LLM Routing},
  author       = {Tao Feng and Haozhen Zhang and Zijie Lei and Haodong Yue and Chongshan Lin and Jiaxuan You},
  year         = {2025},
  howpublished = {\url{https://github.com/ulab-uiuc/LLMRouter}},
  note         = {GitHub repository}
}

@article{zhao2025majority,
  title={The majority is not always right: Rl training for solution aggregation},
  author={Zhao, Wenting and Aggarwal, Pranjal and Saha, Swarnadeep and Celikyilmaz, Asli and Weston, Jason and Kulikov, Ilia},
  journal={arXiv preprint arXiv:2509.06870},
  year={2025}
}

@inproceedings{zheng2024weak,
  title={Weak-to-strong extrapolation expedites alignment},
  author={Zheng, Chujie and Wang, Ziqi and Ji, Heng and Huang, Minlie and Peng, Nanyun},
  booktitle={ICML 2024 Workshop on Models of Human Feedback for AI Alignment},
  year={2024}
}
\bibliographystyle{icml2026}

\newpage
\appendix
\onecolumn

\begin{wraptable}{r}{0.5\textwidth}
\centering
\scriptsize
\setlength{\tabcolsep}{3pt}
\renewcommand{\arraystretch}{1}
\resizebox{1\linewidth}{!}{
\begin{tabular}{lccc}
\toprule[1.5pt]
\multirow{2}{*}{Dataset} &\multirow{2}{*}{Source} &\multicolumn{2}{c}{Size} \\\cmidrule{3-4}
& &dev &test \\\midrule
AGIEval & \citep{zhong2024agieval} & 1156 & 1156 \\
ARC-challenge & \citep{clark2018think} & 299 & 1172 \\
MMLU-redux & \citep{gema2025we} & 1000 & 1000 \\
BBH & \citep{suzgun2023challenging} & 1000 & 1000 \\
GSM8k & \citep{cobbe2021training} & 200 & 1000 \\
MATH & \citep{hendrycks2020measuring} & 956 & 956 \\
GPQA-diamond & \citep{rein2024gpqa} & 99 & 99 \\
TableMWP & \citep{ludynamic} & 124 & 125 \\
TheoremQA & \citep{chen2023theoremqa} & 400 & 400 \\
WikiDYK & \citep{zhang2025bidirectional} & 1000 & 765 \\
PopQA & \citep{mallen2023not} & 1000 & 1000 \\
BLEND & \citep{myung2024blend} & 1000 & 1000 \\
TruthfulQA & \citep{lin2022truthfulqa} & 200 & 617 \\
Sciriff & \citep{wadden2025sciriff} & 114 & 139 \\
Human Interest & \citep{feng2025model} & 400 & 400 \\
\bottomrule[1.5pt]
\end{tabular}
}
\vspace*{10pt}
\caption{Statistics of employed datasets.}
\label{tab:dataset_statistics}
\end{wraptable}

\section{Limitations}
\vspace*{10pt}
Bounded by computational budgets, we execute the single-multi evolution loop for a maximum of three iterations. We envision future experiments on further scaling the single-multi evolution loop: more models, more diverse models, more evolution iterations, would all have potential in yielding stronger models and recipes.

Model collaboration is an extensive domain of research featuring diverse collaboration algorithms and strategies. Bounded by computational budgets, we were able to experiment with seven model collaboration algorithms in total, spanning API-level, text-level, logit-level, and weight-level collaboration strategies. We envision future work on investigating which types and which specific collaboration algorithms are most compatible with the single-multi evolution loop.

We employed four model pool settings in total in the experiments: specialized LMs, general-purpose LMs, the official Qwen variants, and a pool of models with varying sizes. Other model pool settings are possible too: general-specialized collaboration, the collaboration of post-trained aligned models and pretrained base models, etc. We envision future work on expanding the scope of participating models in the single-multi evolution loop.

\vspace*{10pt}
\section{Experiment Details}

\paragraph{Dataset Details} We employ 15 evaluation datasets spanning six domains to evaluate the single-multi evolution loop. We present dataset statistics in Table \ref{tab:dataset_statistics}. For every dataset and model pool, there is at least one (distillation, collaboration algorithm) setting where the improvements to single models and multi-model systems are statistically significant with $p<0.05$ using one-tailed z-test.

\paragraph{Hyperparameter Details}
\begin{itemize}[leftmargin=*] 

\item For trained router, we employ the \textsc{Qwen-2.5-7B} model to initialize the router, employ the \textsc{Skywork/Skywork-Reward-Llama-3.1-8B-v0.2} reward model \citep{liu2024skywork} to derive routing labels.

\item For multiagent debate, we let the models debate and refine answers for 3 iterations before using the best model on the dev set to summarize the final responses.

\item For logit fusion, we assign an equal weight to the next-token probability distributions across all models.

\item For dare-ties model merging, we employ the implementation at MergeKit \citep{goddard-etal-2024-arcees} and use the default hyperparameters there.

\item For LLM blender, we employ \textsc{Qwen-2.5-7B} to initialize the ranker and fuser, and employ top-k=3 for re-ranking and filtering.

\item For multiagent finetuning, we employ three rounds of training, $1e-5$ learning rate, and an effective batch size of 16.

\item For graph routing, we employ the \textsc{sentence-transformers/all-MiniLM-L6-v2} embedding model.

\item For sparta alignment, we employ three iterations of training, 500 instructions per competition batch, and $1e-5$ learning rate for DPO training.

\end{itemize}

\end{document}